\pdfoutput=1
\documentclass[11pt]{article}
\usepackage[]{acl}

\usepackage{silence}
\usepackage[utf8]{inputenc}
\usepackage[T1]{fontenc}
\usepackage{newtxtext}
\usepackage{newtxmath}
\usepackage{inconsolata}
\usepackage{microtype}

\usepackage{graphicx}
\usepackage[table,dvipsnames]{xcolor}
\usepackage{subcaption}
\usepackage{algorithm}

\usepackage{booktabs}
\usepackage{longtable}
\usepackage{makecell}
\usepackage{multirow}
\usepackage{tabularx}

\usepackage{algpseudocode}
\usepackage{tcolorbox}
\usepackage{enumitem}
\usepackage{amsmath}

\usepackage{amssymb}
\usepackage{physics}
\usepackage{bm}
\tcbuselibrary{skins}
\usepackage{cleveref}
\crefname{algorithm}{algorithm}{algorithms}
\usepackage[size=tiny]{todonotes}

\definecolor{rq1color}{RGB}{200,30,30}    
\definecolor{rq2color}{RGB}{30,120,200}   
\definecolor{rq3color}{RGB}{20,150,100}   
\definecolor{rq4color}{RGB}{180,100,0}

\newcommand{\compr}[1]{$\textit{Complex}_\textit{{ratio}}$}
\newcommand{\adl}[1]{\textit{ADL}}
\newcommand{\scv}[1]{\textit{SC}}

\newtcolorbox{takeaway}{
  enhanced,
  colback=blue!5!white, 
  colframe=blue!75!black, 
  coltitle=blue!90!black, 
  fonttitle=\bfseries,
  title=Take-away,
  sharp corners,
  borderline west={3pt}{0pt}{blue!75!black},
  attach title to upper,
  after title={:\enskip},
  colupper=black 
}

\title{%
    Syntactic Complexity Convergence in Dialogue:\\ Analysis of the Phenomenon and Application to LLM Detection
}

\author{%
 \textbf{Yu Wang\textsuperscript{1, 2}} \:
 \textbf{Yanran Chen\textsuperscript{3}} \:
 \textbf{Yifan Wang\textsuperscript{4}} \:\\
 \textbf{Steffen Eger\textsuperscript{3}} \:
 \textbf{Hendrik Buschmeier\textsuperscript{1}}\\
 \textsuperscript{1}Digital Linguistics Lab, Department of Linguistics, Bielefeld University\\
 \textsuperscript{2}Institute for Language Sciences, Utrecht University\\
 \textsuperscript{3}NLLG Lab, University of Technology Nuremberg\\
 \textsuperscript{4}Department of Language Science and Technology, Saarland University\\
}

\begin{document}
\maketitle

\begin{abstract}
    We revisit the under-explored phenomenon of syntactic complexity convergence in dialogue. Extending prior topic-level analyses to full conversations, we examine five human-human datasets and a set of human-LLM dialogues. We integrate and evaluate a multidimensional set of syntactic complexity metrics to quantify structural variation in dialogue. Our analyses reveal robust syntactic complexity convergence across human-human datasets, especially in task-oriented interactions, and statistical tests confirm that these effects are not due to chance. In contrast, human-LLM dialogues show no significant convergence under individual metrics; however, linear-CKA uncovers a distinct cross-metric convergence pattern. Motivated by these differences, we use syntactic complexity as a feature for detecting LLM-involved dialogue, achieving 96\% accuracy. Our classification results demonstrate its potential without relying on likelihood-based signals such as surprisal, which require a language model as the estimator.
\end{abstract}

\section{Introduction}
\label{sec:introduction}

In dialogue, mutual understanding is achieved through adaptive processes that reduce the communicative effort of participants and facilitate successful interaction \cite{pickering2004interactive, pickering2006alignment, reitter-moore-2007-predicting, reitter2014alignment}. Extensive research has shown that humans adjust their language use to align with their conversation partners (e.g., \citealp[]{reitter2014alignment, BuschmeierKopp2018.AAMAS, giulianelli-fernandez-2021-analysing}). This co-adaptivity in interlocutors’ verbal behaviour manifests in linguistic representations across multiple levels of processing \citep[p.~176]{pickering2004interactive}. As interlocutors develop mutual understanding, their linguistic patterns tend to align. Such convergence has been measured at a more abstract level through shifts in syntactic complexity \cite{xu2016convergence} and semantic similarity \cite{xu-2021-global}.
While lexical and semantic alignment are well-documented, the specific mechanism of syntactic complexity convergence, where interlocutors mirror the structural density of each other's language, remains less understood in dialogue. Previous research on complexity matching has primarily used small datasets and mainly focused on short linguistic units, e.g., the so called T-unit, the shortest grammatically allowable sentence where writing can be split \cite{hunt1970syntactic}, or topic episodes \cite{xu2016convergence}. Thus, it remains unclear whether these patterns hold across entire conversations or across different languages. The rapid integration of Large Language Models (LLMs) into everyday communication also raises a new question: Do human–LLM conversations show complexity matching patterns similar to those in human–human dyads? In this paper, we revisit syntactic complexity convergence by analysing a diverse corpus of natural and LLM-involved dialogues. We address gaps in the literature by examining the phenomenon across conversations rather than short linguistic units.
Our contributions are:

\begin{itemize}

    \item \textbf{A multidimensional metric and large-scale multilingual analysis of syntactic complexity:}
    We integrate a comprehensive metric that captures syntactic complexity from multiple structural perspectives. We verify its efficacy by conducting inter-correlation analysis on the metrics and using an open parallel corpus containing sentence pairs controlled for syntactic complexity variation, and apply it to five human–human dialogue datasets across three languages (English, German, Japanese) to examine syntactic complexity convergence on a large scale.

    \item \textbf{A comparison of syntactic complexity in human–human and human–LLM dialogue:}
    We analyse the difference between human–human and human–LLM dialogue from the perspective of syntactic complexity. The analysis provides the basis for an LLM-involved dialogue detection task.\footnote{%
        We consider this to be our novel point: to the best of our knowledge, the most relevant prior work, \citet{bhatt-rios-2021-detecting}, is a chatbot-detection study focused on dialogues generated by pre-LLM language models. More recent LLM-era studies such as DetectGPT \cite{mitchell2023detectgpt} and Fast-DetectGPT \cite{bao2024fastdetectgpt} address LLM-generated text detection in general, but not specifically dialogue.
    }
    
\end{itemize}

\begin{figure}
    \centering
    \includegraphics[width=1\linewidth]{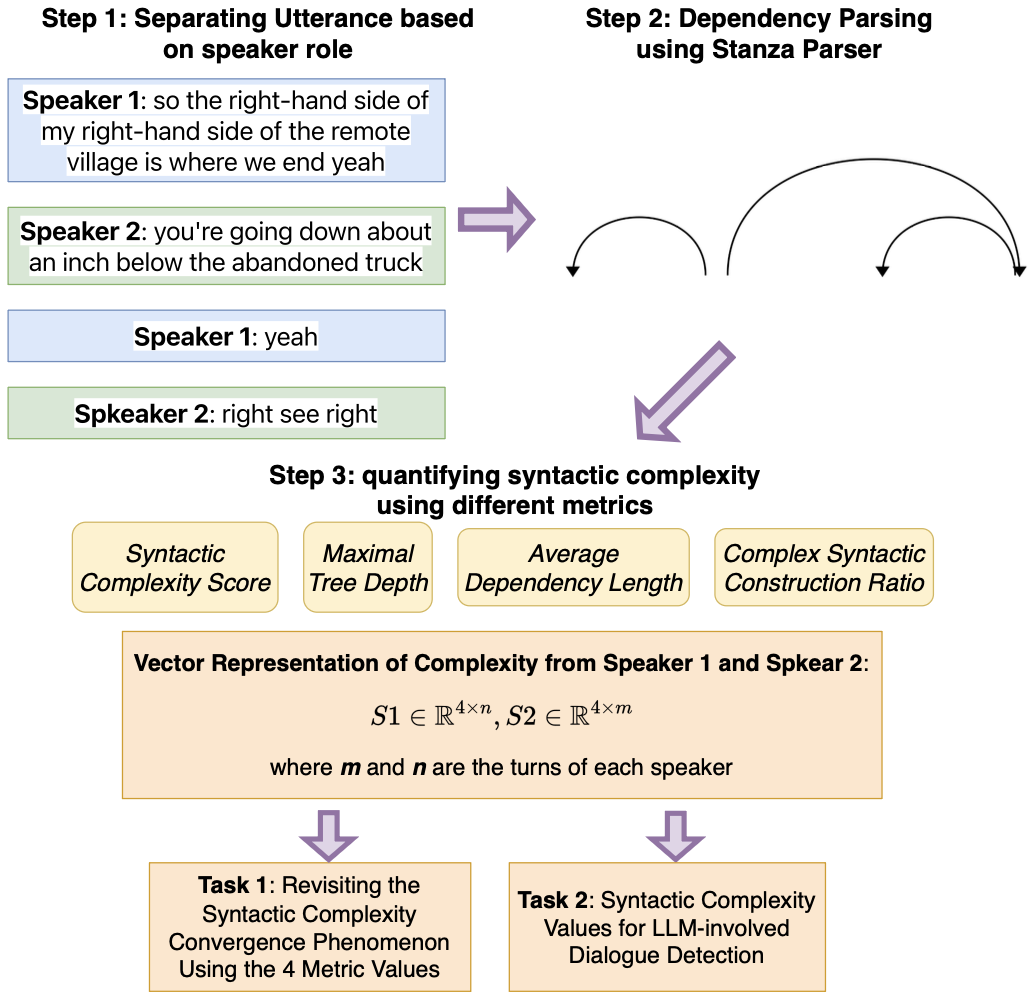}
    \caption{%
        Workflow for the syntactic complexity metrics (example dialogue taken from the Map Task dataset.}
    \label{fig:workflow}
\end{figure}

\vspace{-.2cm}
\section{Background}
\label{sec:background}

\subsection{Complexity Matching as a Type of Linguistic Alignment}

The concept of `complexity matching' was first studied by \citet{abney2014complexity}, who analysed speech signals from conversational speech in terms of time series of acoustic onset events. As a type of abstract behavioural matching, it was later further tested by \citet{xu2016convergence} and \citet{xu-2021-global} who found evidence for complexity convergence on the syntactic and the semantic level. Matching of complexity in dialogue is theoretically grounded in the \textit{Interactive Alignment Model} \citep{pickering2004interactive}, which posits that linguistic alignment facilitates mutual understanding.

While previous research has primarily focused on the development of linguistic alignment in human interaction, recent studies have begun to explore these dynamics within Large Language Models.
For example, \citet{frisch-giulianelli-2024-llm} show that certain types of persona-conditioned LLMs can be adapted to exhibit linguistic alignment toward their conversational partner. Further, clear evidence is provided by \citet{molnar-etal-2023-attribution}, whose empirical experiments uncover that LLMs exhibit language use behaviour (lexical repetition in the interaction) that is similar to humans.
Moreover, \citet{kandra-etal-2025-llms} demonstrate that two LLM agents acting as conversation partners, exhibit mutual syntactic adaptation. Extending this to different setting, \citet{blevins2025language} investigate convergence through dialogue completion tasks, finding that syntactic accommodation (measured mainly through utterance length) is significantly more pronounced in human–LLM interactions than in human baselines. Building on this previous work, we investigate syntactic complexity convergence as a high-level behavioural phenomenon and focus on the change in syntactic complexity between dialogue participants across interactional contexts. By analysing dialogues ranging from human–human to human–LLM dyads, our study aims to characterize how syntactic complexity, as a potential linguistic feature, is negotiated and aligned in both natural dialogue as well as when one interlocutor is an LLM-based conversational agent. If obvious differences exist in syntactic complexity between human--human and human--LLM dialogue, we will discuss the feasibility of using these differences to detect LLM-involved dialogue.

\subsection{Syntactic Complexity Quantification}

Syntactic complexity is a formal linguistic property associated with increased cognitive load during parsing \citep{szmrecsanyi2004operationalizing, lin1996structural}. Researchers typically evaluate syntactic complexity through three lenses: \textbf{word count} (length as a proxy for structural volume), \textbf{node count} (phrasal density), and the \textbf{Index of Syntactic Complexity}, which tracks clausal subordination and morphological embeddedness \citep{beaman1984coordination, givon1991markedness}.
To quantify syntactic complexity, \citet{xu2016convergence} utilize three structural measures:
(i) \textbf{Sentence Length}, a proxy for structural size;
(ii) \textbf{Tree Depth}, the maximum vertical path in the parse tree, where deeper trees indicate greater syntactic complexity; and
(iii) \textbf{Branching Factor}, the average number of children per non-leaf node, reflecting constituent density within a sentence.
The quantification methods proposed in \citet{xu2016convergence} were later adapted and used in follow-up work. \citet{chen2024syntactic}, for example, look at syntactic features, such as tree depth, to analyse the syntactic change in German, and \citet{bagdasarov-alves:2025:LM4DH} use complexity quantification methods to build a syntax-aware classifier for detecting LLM-generated text.

\section{Syntactic Complexity Metrics}
\label{sec:syncom_metric}

\begin{figure}
\includegraphics[width=1\linewidth]{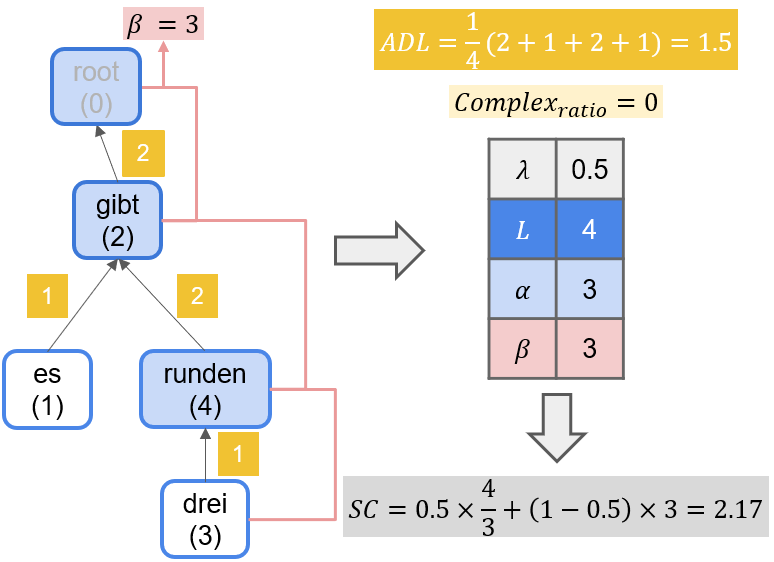}
\caption{%
Illustration of calculating/observing $\beta$, \textit{ADL}, \textit{SC}, $\textit{Complex}_\textit{ratio}$ for the German example sentence “Es gibt drei Runden” (\textit{There are three rounds}). Values in parentheses denote token position indices. Edge labels indicate the dependency distance for each dependency.}
\label{fig:measure_example_short}
\vspace{-1em}
\end{figure}

In this section, we introduce four metrics for quantifying syntactic complexity of utterances from different perspectives, based on prior work. For each metric, larger values indicate greater syntactic complexity. As an example, we illustrate how four metrics are computed using the German sentence “Es gibt drei Runden.” (\textit{there are three rounds}; see Figure~\ref{fig:measure_example_short}).
All metrics are derived from dependency parsing \citep{nivre2010dp}, which identifies grammatical relations between tokens in a sentence by assigning each dependent token a head and a relation label. The resulting structure forms a directed tree (Figure~\ref{fig:measure_example_short}, left), where each dependent points to its head and the root token is attached to a pseudo-root. In this work, we use \href{https://stanfordnlp.github.io/stanza/depparse.html}{Stanza} \citep{qi2018universal} from the StanfordNLP toolkit to segment utterances into sentences and obtain their dependency parses.

Human dialogue data often contain syntactically incomplete utterances, disfluencies (e.g., self-repairs or repetitions), and discourse markers such as fillers (“uh”, “um”) or backchannels (“uh-huh”). We choose Stanza because it can parse incomplete or grammatically irregular sentences and treats such discourse markers as, e.g, modifier tags. Based on the resulting dependency structures, we compute syntactic complexity using the following measures:

\paragraph{Maximal Tree Depth ($\beta$)}
Tree depth is widely used to measure syntactic complexity \citep[e.g.,][]{xu2016convergence,chen2024syntactic}. The depth of a node is defined as its distance from the root, measured by the number of edges on the path between them. The maximal tree depth is the largest node depth in the tree. In Figure~\ref{fig:measure_example_short}, the maximal tree depth ($\beta = 3$) occurs along the path (root) $\rightarrow$ \textit{gibt} $\rightarrow$ \textit{runden} $\rightarrow$ \textit{drei}. The algorithm for maximal tree depth calculation is provided in Appendix~\ref{ap:alg1}.

\paragraph{Average Dependency Length (\textit{ADL})}
Dependency distance is the linear distance between a head and its dependent, calculated as the absolute difference between their position indices. The average dependency length is commonly used as an indicator of syntactic complexity \citep{liu2008dependency, futrell2015dependency}. For example, the distance between (\textit{gibt}, \textit{es}) in Figure~\ref{fig:measure_example_short} is $1$. We compute this distance for each dependency pair and average over all dependencies to obtain a sentence-level complexity score:
\begin{equation*}
\mathrm{ADL} = \frac{1}{|D|} \sum_{(h, d) \in D} |h - d|,
\end{equation*}
where $h$ and $d$ denote the position indices of the head and dependent tokens, and $D$ is the set of dependencies in the sentence. In the example in Figure~\ref{fig:measure_example_short}, $\mathrm{ADL} = 1.5$.

\paragraph{Quantified Syntactic Complexity Value (\textit{SC})}

We compute the syntactic complexity \scv{} of a sentence following \citet{wang-buschmeier-2024-revisiting}:
\begin{equation*}
    \mathit{SC} = 
    \begin{cases}
        \lambda \cdot \frac{L}{\alpha} + (1 - \lambda) \cdot \beta  
            & \text{if } \alpha > 0,\\
        (1 - \lambda) \cdot \beta  
            & \text{otherwise.}
    \end{cases}
\end{equation*}

Here, $L$ denotes sentence length, $\alpha$ the number of heads, $\beta$ the maximal tree depth, and $\lambda$ a tuning factor (set to $0.5$).
The formula combines the effects of sentence length and syntactic structure. Longer sentences tend to have higher syntactic complexity, often resulting in deeper parse trees. The term $\lambda \cdot \frac{L}{\alpha}$ captures how many constituents are dominated by a head, similar to the branching factor proposed by \citet{xu2016convergence}. For the example sentence in Figure~\ref{fig:measure_example_short}, $L = 4$, $\alpha = 3$ (including the root), and $\beta = 3$, resulting in $\mathit{SC} = 2.17$.

\paragraph{Complex Dependency Structure Ratio ($\textit{Complex}_\textit{{ratio}}$)}

In addition to sentence length, syntactic complexity can also be estimated by the proportion of complex structures \citep{lu2010automatic}. We compute \compr{} as the proportion of \emph{long} dependency edges, defined as edges with distance $>3$. Such long dependencies often reflect embedded or inflected constructions. In the example sentence, no dependency edge has a distance greater than 3, so no edge is counted as long and \compr{}$=0$, reflecting the simplicity of the utterance.

\medskip\noindent
Figure~\ref{fig:1df} in Appendix~\ref{ap:data} illustrates this with two dependency trees: one for the simple example sentence (Figure~\ref{fig:measure_example_short}) and another for a more complex sentence. The simple sentence has $\beta = 3$, \textit{ADL} $= 1.5$, \textit{SC} $= 2.17$, and \compr{} $= 0$. In contrast, the complex sentence in Figure~\ref{fig:dpcom} has $\beta = 4$, \textit{ADL} $= 2.428$, \textit{SC} $= 3$, and \compr{} $= 0.48$.\footnote{If an utterance contains multiple sentences, we average their values to represent its syntactic complexity (see Appendix~\ref{ap:data}).}

\subsection{Integrated Complexity Metrics}

When integrating the above four metrics to quantify syntactic complexity, we notice that previous work proposing these individual metrics lack systematic evaluation of their validity in combination. Therefore, we validate our integrated metric framework through three complementary analyses: (i) an inter-correlation analysis confirming that the metrics capture distinct dimensions of syntactic complexity; (ii) a corpus-level evaluation on \textbf{DEplain}, a parallel dataset
of German sentences and their simplified equivalences \cite{stodden-etal-2023-deplain}, confirming directionally correct behaviour; (iii) a qualitative sentence-pair analysis confirming sensitivity to explicit syntactic variation.

For the inter-correlation analysis, we examine whether any single metric sufficiently represents the others. The results reveal consistently weak correlations between several metric pairs, most notably between Tree Depth and Complex Dependency Ratio across datasets, which indicates that each metric contributes unique information and thus all four metrics are necessary to capture syntactic complexity from different structural perspectives (details in Appendix~\ref{ap:deplain} and Figure~\ref{fig:intercor_dialogue}).

For the corpus-level evaluation, we apply our metrics to the \textbf{DEplain} corpus. We acknowledge that simplified sentences in \textbf{DEplain} may undergo lexical or semantic simplification rather than purely structural reduction, which means the corpus provides a conservative test of our metrics' sensitivity to syntactic complexity specifically. Nevertheless, as reported in Table~\ref{tab:syncom_deplain}, all metrics except \compr{} consistently assign lower complexity scores to simplified sentences in both training and test sets, demonstrating directionally correct behaviour at the corpus level.

Moreover, to ensure our metrics are specifically sensitive to syntactic rather than lexical simplification, we additionally sampled a subset of sentence pairs with explicit structural complexity differences for qualitative analysis. As shown in Figure~\ref{fig:sample_syncom_deplain} (Appendix~\ref{ap:deplain}), all four metrics successfully capture the complexity reduction in these cases, further supporting empirical validity of the four metrics for measuring syntactic complexity in dialogue data.

\section{Data}
\label{sec:data}

We run our experiments on multiple human–human dialogue datasets. Some of them are task-oriented dialogues, while others are spontaneous conversations. 
We also incorporate human–LLM dialogues to examine whether LLMs exhibit distinct patterns of syntactic complexity in human–LLM dialogues compared to their human counterparts in human-human dialogues during the development of the conversation. 
A criterion for datasets selection was that dialogues are transcribed such that turns, (anonymous) speakers IDs, and smaller aspects of language use (e.g., stuttered speech, pauses, backchannels) are available. In total, we use five human dialogue datasets, which cover the languages German, Japanese and English:
    \textit{MUNDEX} \citep[ger;][]{TurkWagner2023}, 
    \textit{Hazumi} \citep[jpn;][]{kazunori2021}, 
    \textit{BTSJ} \citep[jpn;][]{usami2023},
    \textit{CANDOR} \citep[eng;][]{reece2023candor}, and
    \textit{Map Task} \citep[eng;][]{anderson1991hcrc}. 
For the human–LLM dialogues, we use the \textit{LMSYS-Chat-1M} dataset \citep{zheng2024lmsyschatm}, from which we select the English and German parts (given their comparatively large proportion among the other language resources).
Detailed descriptions of the datasets, pre-processing, and statistics of the used dialogues can be found in Appendix~\ref{ap:db} and~\ref{ap:data}. 

The dialogue datasets in this study can be categorised into the following three types:
    (i) task-oriented dialogue or dialogue with fixed topics (\textit{MUNDEX}, \textit{MAP TASK}, \textit{Hazumi});
    (ii) everyday conversation dialogue (\textit{BTSJ}, \textit{CANDOR});
    (iii) LLM-involved dialogues (\textit{LMSYS-Chat-1M}).
We define the dialogical roles as ‘initiator’ and ‘follower’ as follows: for the human–human dialogues, the person who initiates the conversation is the initiator and the others are followers; for the human–LLM dialogues, human users are considered initiators and LLM agents followers.
\section{Results and Discussion}
\label{sec:sc_result}

\subsection{Trend of the Syntactic Complexity in Dialogues Based on Mean Value}
\label{subsec:sc_mean}

One of the challenges for tracking the general trend of syntactic complexity convergence in dialogue is that dialogues differ in length, which makes their comparison inconsistent and thus suboptimal. 
A common practice is to divide dialogue into different time steps or relative positions within dialogue (see, e.g., \citealp[]{xu-2021-global, blevins2025language}).
We use a similar solution introduced by \citet{fuscone-etal-2020-contribution}, where dialogues are split up into multiple phases with each phase containing the same number of turns, disregarding the variation in length of dialogues when verifying convergence. 
As we consider convergence as a phenomenon in dialogue which can only be observed if the dialogue contains enough turn-shifts, we set the number of phases to 8 in order to provide enough granularity for checking the general trend.\footnote{%
     Dividing each dialogue into eight phases also means that a dialogue can only qualify for our evaluation if both speakers have more than eight turns (see Appendix~\ref{ap:data} for further explanation on setting the number of phases to 8).
}

Figures~\ref{fig:mean_trend} and~\ref{fig:mean_trend_sup} in Appendix~\ref{ap:sup_fig} together plot the mean values of our metrics across 8 dialogue phases in the datasets. For each phase in each qualified dialogue, we have four time series of quantified syntactic complexity from every conversation participants, and we calculate the average value on each time series to represent the syntactic complexity for that phase for each dialogue role (either initiator or follower). 
Among the five human dialogue datasets, \textit{MUNDEX} and \textit{Hazumi} exhibit the most pronounced convergence (Figure~\ref{fig:mean_trend}): in the \textit{MUNDEX} dataset, the syntactic complexity of initiators gradually decreases and converges with that of followers. This tendency is clearly evident in the first three metrics. In contrast to that, in the \textit{Hazumi} dataset, the syntactic complexity from initiators increases and adapts to the equivalences from the followers. This tendency is clearly evident in the first three metrics as well. While \textit{BTSJ} and \textit{Candor} (Figures~\ref{fig:mean_trend_sup}a and Figure~\ref{fig:mean_trend_sup}b in Appendix~\ref{ap:db}) also display a discernible tendency toward convergence, the \textit{Map Task} dataset as human--human dialogue dataset shows no observable trend of syntactic complexity convergence, and similarly, both human--LLM dialogue datasets have no observable trend of syntactic complexity convergence (Figure~\ref{fig:mean_trend_sup}c to \ref{fig:mean_trend_sup}e in Appendix~\ref{ap:db}).

With regard to the reason why \textit{MUNDEX} and \textit{Hazumi} exhibit more pronounced convergence, we have taken a closer look into the dialogues and came up with the following hypothesis: in both dialogue datasets, one of the speakers may have better topic-related knowledge compared to the others. In case of the MUNDEX corpus, where explainers, as dialogue initiators, explain a board game to the explainees, explainers tend to adjust their explaining strategies, i.e., simplifying their utterances, to make the opponents pick up the game rules more easily. In response, explainees start to initiate more complex questions (as shown in the plot, a peak in phase 6). For the Hazumi corpus, the dialogue initiators, who are familiar with the topics, initiate the talk on one topic with the followers and then expand the conversation based on the followers' utterances. In response, the dialogue initiators' utterances become more syntactically complex.

\smallskip\noindent
\colorbox{gray!15}{%
  \parbox{0.98\linewidth}{\textbf{Takeaway}: Through visualisation, we notice that syntactic complexity convergence appears to be more obvious in human--human dialogue than human--LLM dialogue.}}

\begin{figure*}[t!]
    \begin{subfigure}[b]{1\textwidth}
        \includegraphics[width=\textwidth]{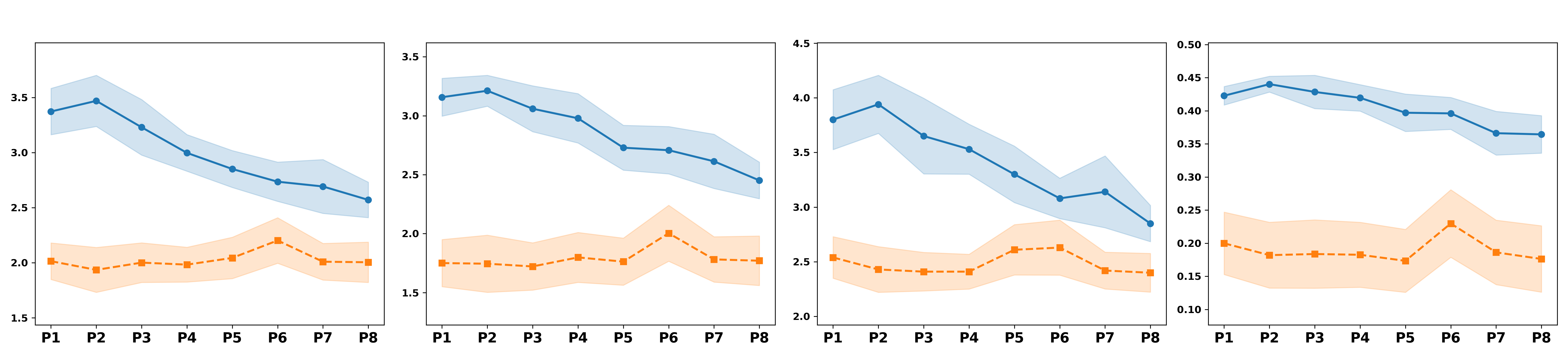}
        \caption{MUNDEX corpus (ger)}
    \end{subfigure}
    \begin{subfigure}[b]{1\textwidth}
        \includegraphics[width=\textwidth]{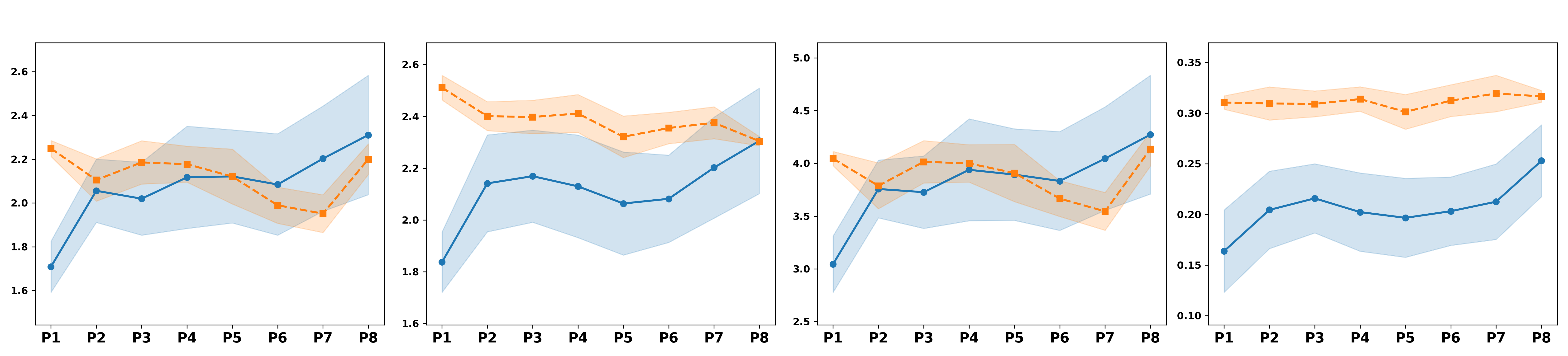}
        \caption{Hazumi Corpus (jpn)}
    \end{subfigure}
    \caption{%
        The plots show the syntactic complexity across phases for \textit{MUNDEX} and \textit{Hazumi}. The \textbf{blue line is the Dialogue Initiator}, the \textbf{orange line is the Dialogue Follower}. The four metrics in each sub-figure from left to right are: \textbf{syntactic complexity value}, \textbf{dependency length}, \textbf{tree depth}, and \textbf{complex Dependency Structure Ratio}. We calculate the mean value based on the eight phases (P1-P8) we pre-defined with $95\%$ confidence interval.} 
    \label{fig:mean_trend}
\end{figure*}

\subsection{Convergence Patterns and Ratio}
\label{sec:ratio_pattern}
As discussed in Section~\ref{subsec:sc_mean}, when we take the average value of quantified syntactic complexity on each pre-define phase, syntactic complexity convergence can be observed in different natural dialogue datasets. However, more evidence is still needed in order to ensure the validity of our observation. Our next goal is to check, to what degree the phenomenon of syntactic complexity convergence can be confirmed. Specifically, we want to analyse (i) the patterns of convergence, and (ii) the concrete ratio of convergence in each dataset. 

To verify the ratio of convergence as well as the convergence patterns, we need to calculate the complexity difference between the initiator and follower along the 8 different phases. First, in order to verify that convergence occurs, in each selected dialogue, we check if the difference between the two speakers decreases between the first and the last phase: 
    $|P^1_{\mathit{init}} - P^1_{\mathit{fol}}| > |P^8_{\mathit{init}} - P^8_{\mathit{fol}}|$.
The second constraint requires the sequence of differences to show a consistent downward trend, with at least two decreases across phases. We consider two convergence patterns. In Pattern 1, the syntactic complexity of both speakers decreases and converges to a lower value, which may occur when a conversation ends and speakers produce simpler utterances. In Pattern 2, the speakers converge to a point where one speaker’s syntactic complexity decreases while the other’s increases. \textit{If any of the four complexity metrics shows such convergence with statistical significance, we treat it as evidence of syntactic complexity convergence in the dataset.} Based on this criterion, Pattern 2 appears more frequently than Pattern 1 (see Table~\ref{tab:pattern_ratio_summary} in the Appendix).

We then report the proportion of dialogues in each dataset that exhibit syntactic complexity convergence. To ensure that the ratios are not due to random effects, we generate null distributions by randomly shuffling the turns within each dialogue $B$ times (scaled by dataset size; e.g., 1500 permutations for large datasets such as \textit{BTSJ}, \textit{CANDOR}, and \textit{LLM\_English}, and 1000 for smaller datasets such as \textit{Hazumi} and \textit{MUNDEX}). This procedure breaks the temporal structure while preserving turn and role information. If the values from the shuffled data are similar to or larger than the observed values, we conclude that convergence is not present. 

The results of statistical testing, as summarized in Table~\ref{tab:convergence_results} where $p$-values as well as Cohen's $d$ with 95\% confidence intervals are reported, show that human task-oriented dialogue datasets exhibit statistically significant syntactic complexity convergence, whereas the human--LLM datasets do not.

\smallskip\noindent
\colorbox{gray!15}{%
  \parbox{0.98\linewidth}{\textbf{Takeaway}:     Quantitative analysis shows that \textbf{convergence pattern 2} is dominant in all of the chosen datasets. Under the eight phases setting, syntactic complexity convergence can be statistically confirmed in four out of five human-human dialogue datasets, but not in the human-LLM dialogue datasets.}}

\begin{table*}[t]
\scriptsize
\setlength{\tabcolsep}{5pt}
\centering
\begin{tabular}{llllccccc}
\toprule
\textbf{Dataset ($N$ Dialogues)} & \textbf{Dialogue Type} & \textbf{Metric} & \textbf{Observed} & \textbf{Random (Mean)} & \textbf{$p$-value} & \textbf{Cohen's $d$} & \textbf{95\% CI} \\
\midrule

MUNDEX (50) & Task-oriented / Fixed topic & Syn. value & \textbf{0.7600$^{***}$} & 0.4775 & <0.001 & \textbf{4.150} & \textbf{[0.1891, 0.3743]} \\
 & & Dependency length & \textbf{0.6600$^{**}$} & 0.4819 & 0.009 & \textbf{2.538} & \textbf{[0.0848, 0.2757]} \\
 & & Tree depth & \textbf{0.5600$^{***}$} & 0.2657 & <0.001 & \textbf{4.962} & \textbf{[0.1889, 0.4100]} \\
 & & Complex dep. ratio & \textbf{0.6000$^{*}$} & 0.4662 & 0.032 & \textbf{1.899} & \textbf{[0.0338, 0.2375]} \\
\midrule

Hazumi (33) & Task-oriented / Fixed topic & Syn. value & 0.5152 & 0.4864 & 0.438 & 0.347 & [-0.0951, 0.1636] \\
 & & Dependency length & \textbf{0.7576$^{**}$} & 0.4937 & 0.002 & \textbf{3.036} & \textbf{[0.1563, 0.3758]} \\
 & & Tree depth & 0.3333 & 0.3723 & 0.750 & -0.490 & [-0.1617, 0.0823] \\
 & & Complex dep. ratio & \textbf{0.6970$^{**}$} & 0.4942 & 0.010 & \textbf{2.376} & \textbf{[0.0940, 0.3392]} \\
\midrule

BTSJ (498) & Everyday conversation & Syn. value & 0.4137 & 0.4705 & 0.997 & -2.593 & [-0.0900, -0.0264] \\
 & & Dependency length & 0.4900 & 0.4873 & 0.473 & 0.116 & [-0.0315, 0.0380] \\
 & & Tree depth & 0.2691 & 0.2928 & 0.881 & -1.189 & [-0.0540, 0.0049] \\
 & & Complex dep. ratio & \textbf{0.5321$^{*}$} & 0.4917 & 0.036 & \textbf{1.800} & \textbf{[0.0034, 0.0728]} \\
\midrule

Candor (1855) & Everyday conversation & Syn. value & 0.3871 & 0.4817 & 1.000 & -8.381 & [-0.1113, -0.0773] \\
 & & Dependency length & 0.4388 & 0.4828 & 0.999 & -3.853 & [-0.0614, -0.0274] \\
 & & Tree depth & 0.2194 & 0.2103 & 0.161 & 1.024 & [-0.0055, 0.0233] \\
 & & Complex dep. ratio & \textbf{0.4642$^{\dagger}$} & 0.4467 & 0.066 & \textbf{1.527} & \textbf{[0.0004, 0.0334]} \\
\midrule

Map Task (121) & Task-oriented / Fixed topic & Syn. value & 0.4380 & 0.4875 & 0.899 & -1.138 & [-0.1140, 0.0188] \\
 & & Dependency length & 0.4215 & 0.4896 & 0.945 & -1.534 & [-0.1306, 0.0037] \\
 & & Tree depth & 0.3471 & 0.3446 & 0.503 & 0.057 & [-0.0625, 0.0688] \\
 & & Complex dep. ratio & 0.3967 & 0.4515 & 0.916 & -1.237 & [-0.1182, 0.0110] \\
\midrule

LLM\_German (234) & LLM-involved & Syn. value & 0.4316 & 0.4822 & 0.945 & -1.553 & [-0.0971, -0.0025] \\
 & & Dependency length & 0.3590 & 0.4818 & 1.000 & -3.782 & [-0.1715, -0.0749] \\
 & & Tree depth & 0.3718 & 0.3826 & 0.659 & -0.345 & [-0.0626, 0.0347] \\
 & & Complex dep. ratio & 0.4658 & 0.4723 & 0.601 & -0.203 & [-0.0534, 0.0443] \\
\midrule

LLM\_English (7850) & LLM-involved & Syn. value & 0.4122 & 0.4726 & 1.000 & -11.013 & [-0.0681, -0.0516] \\
 & & Dependency length & 0.4152 & 0.4722 & 1.000 & -10.410 & [-0.0657, -0.0483] \\
 & & Tree depth & 0.3566 & 0.3789 & 1.000 & -4.128 & [-0.0304, -0.0147] \\
 & & Complex dep. ratio & 0.4511 & 0.4596 & 0.935 & -1.521 & [-0.0158, -0.0004] \\
\bottomrule
\end{tabular}
\caption{Observed dialogue initiator--follower convergence compared to random baseline (8-phase analysis). $N$ represents the number of dialogues meeting the $\geq 8$ turns per speaker requirement. Bold indicates statistical significance ($p < 0.1$). Significance markers: $^{\dagger} p < 0.1$, $^{*} p < 0.05$, $^{**} p < 0.01$, $^{***} p < 0.001$. CIs refer to the 95\% confidence interval for Cohen's $d$.}
\label{tab:convergence_results}
\end{table*}

\subsection{Potential Convergence among Different Metric Values}
\label{sec:cka_conv}

Furthermore, we hypothesize that syntactic complexity convergence is not univariate\footnote{%
    Based on the Interactive Alignment Model that priming can operate across representational dimensions, and evidence that although sharing similarities, priming in LLMs has differences from priming in humans \cite{sinclair-etal-2022-structural, jumelet-etal-2024-language}, we hypothesize that human–LLM convergence may manifest cross-metrically rather than within individual metrics. See Appendix~\ref{ap:data} for the full theoretical motivation.
} but rather multivariate given that: 
    (i) the change of the average dependency length from speaker A can be aligned to the change of the tree depth from speaker B, and therefore the convergence may happen between two different metric values; 
    (ii) the syntactic complexity from speaker A may not be strictly time-aligned with the equivalence from speaker B.
To capture such dynamics, we consider the syntactic complexity of each phase as a matrix representation of the syntactic complexity: for each dialogue phase $p \in \{1, \dots, n\}$, we construct complexity matrices $X_p$ and $Y_p \in \mathbb{R}^{T \times 4}$, where $T$ is the number of turns in the phase and $4$ represents the integrated syntactic complexity metrics. We then calculate the alignment distance $d_p = d(X_p, Y_p)$ to track the convergence between the dialogue initiator and follower. We chose to use the Linear Centered Kernel Alignment \citep[Linear-CKA;][]{kornblith19a}, a distance function that was originally used to compare the similarity of different layers in language models' hidden representation (vector matrix) with the potential that it can address points (i) and (ii) mentioned above. 
Linear-CKA measures the representation distance by the similarity between representations' inter-sample similarity $\langle X_i^{p\top} X_i^p, Y_j^{p\top} Y_j^p \rangle$ in phase $p$ from dialogue speaker $i$ and $j$. Formally, for each dialogue phase $p \in \{1, \dots, n\}$, we construct complexity matrices $X_i^{p}$ and $Y_j^{p} \in \mathbb{R}^{T \times 4}$. We then calculate the alignment distance using:
\begin{equation*}
    d_{\text{CKA}}(X^p_i, Y^p_j) = 1 - \frac{\| X^{p\top}_i Y^p_j \|_F^2}{\| X^{p\top}_i X^p_i \|_F \| Y^{p\top}_j Y^p_j \|_F}
\end{equation*}
where $||\cdot||_F$ denotes the Frobenius norm. 
Before the distance calculation, we first normalise the matrix values by applying MinMax normalisation. The smaller the distance value, the closer the syntactic complexity between dialogue participants. 
In terms of linguistic interpretation, a small linear-CKA-based distance value can suggest that the dialogue participant (human or LLM) is adapting its overall structural complexity profile to match the other. For example, it may be the case that speaker A has less complex syntactic constructions in their utterances, and in response, speaker B produces utterances with larger average dependency length.

After obtaining the distance values from different phases, we fit the data points into a linear regression model to check if a statistically significant descending trend can be observed. 
We, again, set the phase number to eight, making it coherent with the previous experimental setting. As the results reported in Table~\ref{tab:li_cka_8} show, both human-LLM dialogue datasets have a strong tendency of syntactic complexity convergence under cross metric evaluation (negative slope value with statistical significance), which can only be confirmed for one of the human-human dialogue datasets.

\smallskip\noindent
\colorbox{gray!15}{%
  \parbox{0.98\linewidth}{\textbf{Takeaway}: By applying \textbf{Linear-CKA}, we verify a statistically significant tendency of syntactic complexity convergence in Human-LLM dialogue under cross metric setting.}}

\subsection{Discussion}
\label{sec:discussion}

Our first finding is that syntactic complexity convergence is more evident in task-oriented or topic-focused dialogues than in daily conversations. Among the three such datasets (\textit{MUNDEX}, \textit{Hazumi}, \textit{MAP TASK}), two show clear convergence tendencies (Figure~\ref{fig:mean_trend}). In contrast, this tendency is less visible in daily-conversation datasets (Figure~\ref{fig:mean_trend_sup}a and \ref{fig:mean_trend_sup}b in the Appendix), and not observable in \textit{MAPTASK} and \textit{Human--LLM} dialogues (Figure~\ref{fig:mean_trend_sup}c to \ref{fig:mean_trend_sup}e in the Appendix). The statistical analysis in Table~\ref{tab:convergence_results} supports this observation. No statistically significant syntactic complexity convergence is found in LLM-involved dialogues.

As syntactic complexity convergence can be interpreted as a form of linguistic alignment, these findings support the results of \citet{reitter2014alignment}, who report stronger syntactic alignment in task-oriented dialogues than in spontaneous conversations. Furthermore, human––human and human––LLM dialogues show different convergence patterns. Human––human dialogues exhibit convergence when measured within the same metric, whereas human--LLM dialogues do not. By analysing cross-metric convergence using linear-CKA, we see that clear convergence emerges in human–LLM dialogues (details in Table~\ref{tab:li_cka_8}).

We further evaluate the impact of dialogue length by testing different pre-defined lengths in Appendix~\ref{ap:length_effect}. It appears that longer dialogues do not necessarily lead to stronger convergence, except in the \textit{MUNDEX} dataset. 
When shorter dialogues are excluded in the cross-metric analysis, convergence becomes clearer in \textit{MUNDEX} but disappears in \textit{BTSJ} and \textit{LLM\_German}.

\begin{table}
\footnotesize
\centering
\begin{tabularx}{\columnwidth}{lrrrr}
\toprule
    \textbf{Dataset} & \textbf{Slope} & $R^2$ & \textbf{Std.\ Err} & \textbf{$p$-value} \\
\midrule
     \textbf{MUNDEX} & -0.0015 & 0.0142 & 0.0050 & 0.7790 \\
     \textbf{BTSJ} & -0.0046 & 0.9077 & 0.0006 & \textbf{0.0003} \\
     \textbf{Hazumi} & 0.0075 & 0.1701 & 0.0067 & 0.3099 \\
     \textbf{Candor} & -0.0004 & 0.0047 & 0.0026 & 0.8721 \\
     \textbf{Map Task} & -0.0011 & 0.1644 & 0.0010 & 0.3189 \\
     \textbf{LLM (DE)} & -0.0174 & 0.9099 & 0.0022 & \textbf{0.0002} \\
     \textbf{LLM (EN)} & -0.0156 & 0.9480 & 0.0015 & \textbf{<0.0001} \\
\bottomrule
\end{tabularx}
\caption{%
    Illustration of the stacked comparison of 8-phase regression results. As we calculate the linear-CKA-based distance value between the syntactic complexity from both dialogue speakers under the cross-metrics setting, statistically significant negative slope values indicate the occurrence of syntactic complexity convergence.
}
\label{tab:li_cka_8}
\end{table}

\section{Application: LLM Dialogue Detection with Syntactic Complexity Features}
\label{sec:app}

As shown in Section~\ref{sec:sc_result}, (i) human–human dialogues display clearer syntactic complexity convergence within the same metric, whereas (ii) human–LLM dialogues tend to converge across different metrics. These patterns suggest that syntactic complexity may serve as a useful feature for detecting LLM-involved dialogues.

Here, we investigate whether quantified syntactic complexity metric values can be used for human–LLM dialogue detection.\footnote{A related study by \citet{bagdasarov-alves:2025:LM4DH} uses syntactic features such as dependency length and tree depth to detect LLM-generated text. We follow the same idea but apply it to detecting LLM-involved dialogue. \citet{li-etal-2025-prdetect} also report large syntactic-tree differences between human-written and LLM-generated text and build a detector based on this.}
We use our English dialogue datasets (Candor and LLM\_English), whose relatively large sizes make them suitable for classification. The task is to distinguish human–human dialogue (\textit{Candor}) from human–LLM dialogue (\textit{LLM\_English}). Our pipeline builds on \citet{xu-etal-2024-detecting}, who use a Discrete Fourier Transform (DFT) over surprisal values to detect LLM-generated text. Since syntactic complexity metric values also form fluctuating time series and, like surprisal, reflect cognitive load in language processing \cite{hale2001, DEMBERG2008193, demberg-etal-2012-syntactic}, it is reasonable to apply their DFT-based approach here. We therefore conduct DFT over syntactic complexity time series and use the resulting frequency spectra as features for classification.

\begin{table*}
\small
\begin{tabularx}{\textwidth}{llXXXl}
\toprule
    \textbf{Input} & \textbf{Model} & \textbf{F1} & \textbf{Accuracy} & \textbf{AUC} \\
\midrule
    \multirow{3}{*}{Syn Com}
     & Spectrum Length (baseline) & 0.745 (0.074) & 0.753 (0.063) & 0.757 (0.067) \\
     & Linear Regression          & 0.962 (0.042) & 0.960 (0.047) & \textbf{1.000 (0.000)} \\
     & Transformer Encoder        & 0.947 (0.053) & 0.947 (0.053) & 0.993 (0.012) \\
\midrule
    \multirow{3}{*}{Surprisal}
     & Spectrum Length (baseline) & 0.955 (0.012) & 0.956 (0.011) & 0.976 (0.016) \\
     & Linear Regression          & \textbf{0.972 (0.014)} & \textbf{0.972 (0.014)} & 0.997 (0.002) \\
     & Transformer Encoder        & 0.961 (0.017) & 0.960 (0.018) & 0.994 (0.004) \\
\bottomrule
\end{tabularx}
\caption{%
    Classification performance using the FourierGPT pipeline on the saved test set. We compare models using syntactic complexity (“Syn Com”) versus surprisal as input. Reported accuracies are means over 10 runs; standard deviations are shown in parentheses. Bold values indicate the best performance. Additional robustness analyses appear in Appendix \ref{ap:robust} and Table \ref{tab:syncom_fft_filter_results}.}
\label{tab:fft_results}
\end{table*}

We first feed the four syntactic-complexity time series into the FourierGPT pipeline to obtain their spectrum representations. After selecting the top-k peaks (set to 30), we concatenate them to form the feature vectors for classification. As a baseline, we report results using only the spectrum length to train a logistic regression model. In addition to the logistic regression model using spectrum representations, we also train a transformer-based encoder. We compare the original FourierGPT pipeline, which relies on surprisal estimates, with our version using quantified syntactic complexity as the main features, with results shown in Table~\ref{tab:fft_results}. Further technical details and parameter settings are provided in Appendix~\ref{sec:fGPT}.

Using only the spectrum length, surprisal-based features reach 95.5\% F1 score, while syntactic-complexity features reach 74.5\%. This difference stems from the fact that surprisal is computed over words and thus directly reflects dialogue length, whereas syntactic complexity is computed at the utterance level and does not. In our data, human--LLM dialogues compared to human--human dialogues have comparatively shorter length (less turns as shown in Table~\ref{tab:dialogue_statistics}). When using spectral features, surprisal-based inputs still perform better overall (F1 score 97.2\%), but syntactic-complexity features also yield competitive accuracy (F1 score 96.2\%). Unlike surprisal-based detectors, such as \textit{GLTR} \citep{gehrmann-etal-2019-gltr} and \textit{GPT-who} \citep{venkatraman-etal-2024-gpt}, which rely on language models to estimate likelihoods, syntactic-complexity computation requires no language model at all, making it a potentially valuable alternative.

\section{Conclusion}
\label{sec:concluding-remarks}

In this paper, we revisit an under-investigated linguistic phenomenon, syntactic complexity convergence in dialogue, which is considered as an alignment behaviour indicating the development of mutual understanding along the change of the conversation. We extend prior work from topic-level analysis to full dialogues and examine five human–human dialogue datasets, as well as human–LLM dialogues.
We review previous studies and integrate metrics, along with further evaluation, to quantify syntactic complexity. Our empirical analysis shows clear syntactic complexity convergence across the human–human dialogue datasets, particularly in task-oriented settings, and statistical tests confirm that this effect is not random. Although human–LLM dialogues show no significant convergence under the same metrics, linear-CKA reveals a different cross-metric convergence pattern. Given these differences, we use syntactic complexity as a feature for detecting LLM-involved dialogues. Classification results highlight its potential without relying on language-model-based likelihood estimates (e.g., surprisal).

\section*{Limitations}
\label{sec:limitations}

Within the scope of this study, we consider the following four points as potential limitations.

First of all, although our study covers different languages, it appears that in dialogue datasets, English dialogues are still a dominant language resource in comparison to German and Japanese. The same situation holds for human--LLM dialogues. Although \citet{don-yehiya-etal-2025-sharelm} introduce a larger collection of human--LLM dialogues with a more balanced language distribution, we use only \citet{zheng2024lmsyschatm} for human--LLM dialogue in this paper, as metadata details such as language type were missing in \citet{don-yehiya-etal-2025-sharelm} during our last inspection of their dataset.

Secondly, the findings in this paper are highly dependent on the performance of the Stanza parser. As pointed out by \citet{chen2024syntactic}, experimental results on diachronic syntactic change can vary based on which dependency parser is used. That is to say, we can assume that, using a different parser (e.g., Spacy's Dependency Parser, \citealp[]{montani_ines_2023_10009823}), could provide us with slightly different results. 
However, the major reason for not including results from different existing parsers is that, to the best of our knowledge, Stanza is the most reliable off-the-shelf parser that can be used with dialogue data. One of the future works could be to train different parsers that can be used for parsing the dependency relationships of dialogue data and further verify the results reported in this paper.

Thirdly, the phase-based division of dialogues is rather suboptimal. Other elegant solutions can be, e.g., using a segmentation method to divide the dialogues based on, e.g., topic shifts. This, however, requires either a large human annotation effort to train a classifier, or reliable off-the-shelf models to find topic shifts automatically. As we are uncertain about the quality of such models for our dataset, we leave this as future work.

Lastly, for the human-LLM dialogues, we didn't distinguish between different LLMs (GPT-4, LLama, etc) of various sizes. Therefore, in this study we cannot answer research questions such as, `Do LLMs with larger scaling size show stronger tendency of syntactic complexity convergence?', or `Do LLM-A as a dialogue participant show stronger complexity matching to the human speaker compared to LLM-B?'.

\section*{Ethics statement}
\label{sec:ethics}

For all of the selected dialogue datasets, we have checked that there is no revealing of participants' identity. The speaker names are all replaced with unique IDs. 
One of the potential ethical issues we can foresee for the syntactic complexity-based human–LLM dialogue detection is that such an application might be misused. Even if the classification result shows that an LLM is involved in the conversation, it may turn out that human users (e.g., students) are only using LLMs to assist with their writing in the chat. Therefore, such tool should be used to raise awareness that, `LLMs are used to support the conversation/text-based chat'. Any judgement or punishment based on such classification results should be carefully evaluated.

In terms of of our research topic, syntactic complexity convergence as a linguistic phenomenon, requires us to assume that dialogue participants have a mental model of language use in order to adapt to others. However, as noted in \citet{kandra-etal-2025-llms}, LLMs as dialogue participants, are not expected to make conscious decisions to coordinate their conversation with interlocutors. We therefore need to be cautious when dealing with the question of whether LLMs do have a mental model of language use. Although LLMs as dialogue participants, under specific pre-defined evaluation metrics, can show the convergence of syntactic complexity to their communication partner (Section~\ref{sec:cka_conv}), it is still hard to answer if such convergence is more than the outcome of statistical learning or whether the LLM really starts to have a mental model like humans to adapt to its communication partner.

We acknowledge that, writing of the experimental code was supported by ChatGPT and GitHub Copilot. We used Gemini for grammar checking.

\section*{Supplementary Material}
Code and data are available at \url{https://doi.org/10.5281/zenodo.22206634}, or, for your convenience, at \url{https://github.com/owuQQQ/syntactic_complexity_in_dialogue}.

\section*{Acknowledgement}
\label{sec:ack}

The early structure of this paper benefited significantly from insightful discussions with Vera Demberg, particularly regarding the evaluation of syntactic complexity metrics and robustness tests on the convergence phenomenon. We are also grateful for her helpful proofreading of the original draft.
This work is funded by the \href{https://www.dfg.de}{Deutsche For\-schungs\-gemeinschaft (DFG)}: \href{https://gepris.dfg.de/gepris/projekt/438445824}{TRR 318/3 2026 – 438445824}, A02 and B07.
The research reported in this paper was also supported by the European Research Council (ERC), grant 101088098 -- MEMLANG. Views and opinions expressed are however those of the authors only and do not necessarily reflect those of the European Union or the European Research Council Executive Agency. Neither the European Union nor the granting authority can be held responsible for them.
The NLLG Lab gratefully acknowledges support from the German Research Foundation
(DFG) via the Heisenberg Grant EG 375/5-1.
This work also acknowledges the funding from DFG project GRK 2853 ‘Neuroexplicit Models of Language, Vision, and Action’ (project number 471607914).

\bibliography{emnlp}
\bibliographystyle{acl_natbib}

\appendix

\section{Dialogue Dataset Description}
\label{ap:db}
\paragraph{German Dataset:}
\textbf{MUNDEX} is a German dialogue dataset and consists of around 87 task-oriented dialogues and focuses on explanation in interaction \cite{TurkWagner2023}.
Each dialogue is a explanation scenario involving a speaker (the explainer) explaining how to play a board game to a recipient (the explainee).
The dataset in total consists of 87 dialogues between dyads of German native speakers, with each dialogue spans around 15-30 minutes. At its current stage, speech diarization was manually annoted with the help  of Whisper ASR \citep{radford2022robust}. 

\paragraph{Japanese Dataset:} \textbf{Hazumi Dataset} is a collection of 33 Japanese dialogues which are about 10 daily life topics \cite{kazunori2021}. One speaker (virtual agent controlled by human operator) plays the role that when the other shows no evidence of interest toward current conversation topic, the conversation will be guided toward the next topic. The setting is human agent interaction where the agent is a human operator located in a different room using Wizard-of-Oz method, each dialogue is around 15-20 minutes. Given that the agent's utterances are still produced by the human operators based on participants' utterance, we still include this dataset as human-human dialogue dataset. Another dataset is the \textbf{BTSJ} 1000 Person Japanese Natural Conversation Corpus \citep{usami2023}, which consists of 541 spontaneous conversation with no specific pre-defined topics and each dialogue around 5-25 minutes approximately based on our check given that there is no official statistics on that.

\paragraph{English Dataset:} We use the open-sourced \textbf{CANDOR} dataset \cite{reece2023candor}, which is a large-scale collection of nearly 1656 naturalistic, open-ended conversations between strangers through the video chat platform. \textbf{CANDOR} dataset has no specific topic and thus the content is closer to daily spontaneous conversation, average length of each dialogue is about 30 minutes. Additionally, we also include \textbf{Map Task} dataset \cite{anderson1991hcrc}, which is a collection of 128 task-oriented dialogues. Despite its usage for confirming linguistic alignment phenomenon (e.g., \citealp[]{reitter2006priming, reitter2014alignment}), to the best of our knowledge, this corpus is not yet used for verifying the syntactic complexity convergence.

\paragraph{Human-LLM Dialogue dataset:} we used the Human-LLM dialogue data provided by \citet{zheng2024lmsyschatm}, which consists of one million real-world conversations between users and 25 large language models. These `dialogues' \footnote{It is worth noting that, human-LLM conversation is intrinsically different from natural conversation. In natural conversation, for example,interruption and backchanneling are expected to happen from time to time while this is not the case for human-LLM conversation where LLM agents wait human users to finish their turns.} were collected "in the wild" through the Vicuna demo and the Chatbot Arena platform \cite{chatbotArena2023}. We selected the German and English dialogues for our study purpose, given that their sample size is comparatively larger while there is little Japanese data which is qualified for our study.

\section{Dialogue Dataset Pre-processing}
\label{ap:data}

\paragraph{Data Cleaning for Parsing:} Since the calculation of syntactic complexity is mostly based on the syntactic structure of the given utterance, an important requirement for this purpose is that the parser must first be able to recognize the words in the transcribed spoken dialogue. In some of our dialogues we found that some of the transcriptions use informal word forms to reflect the original dialogue content instead of the corresponding correct word formats. This approach keeps the originality of the corpus, while adding difficulty for parsing. Another challenge is that special tokens such as `<p>' and `<smile>' are added to the transcriptions in order to reflect pauses and the paralinguistic feature `<smile>'. This, again, is not processable by the parser. As our solution, when we pre-process such utterances, we replace all incorrect word formats with their correct ones in order to maximally increase the parsing accuracy, which is directly related to the accuracy of the calculation of syntactic complexity. For the special tokens, we remove them.

\paragraph{The Motivation of Setting the Phase Number to 8:} The motivation is firstly based on our review of the widely used dialogue corpus \textit{Switchboard} \cite{switch1992}, which shows that within the 2438 dialogues, the duration of each one is around 3 to 10 minutes, with an average of 6 minutes \cite{swireport1999}, each dialogue covers either 1 or 2 topics (they choose one topic at the beginning and later may shift to another topic). A follow-up study by \citet{soni-etal-2022-empirical} shows that among the 215 selected dialogues from \textit{Switchboard}, an average topic shift was observed to contain 12 turns and the average number of topics per dialogue is around 8; thus, around every 1 minute, a topic shift could happen.

A recent study in the \textbf{CANDOR} corpus proposed that a psychologically meaningful turn is around 12 seconds (Table 1 from \citealp[]{Cooney2025}). If a topic consists of 12 turns, then 12 turns multiplied by 12 seconds is around 2.4 minutes. Moreover, pauses, breathing, and backchanneling are also part of turns which take extra time \cite{SKANTZE2021101178}. Based on all of the above statistics from previous studies, our heuristic estimate for topic shifting is that within 4 minutes, a conversation topic can start and end. Given that the dialogues used in our study range from 15 minutes to 30 minutes in duration, while it is suboptimal, dividing each dialogue into 8 phases can ensure that each phase covers around 1 to 4 minutes. 

\paragraph{The Motivation of Using Linear-CKA in Section~\ref{sec:cka_conv}:} The Interactive Alignment Model \citep{pickering2004interactive} proposes that, linguistic alignment arises through priming, with dialogue participants increasingly sharing lexical items or syntactic structures as the conversation unfolds. Because priming at one linguistic level, can trigger priming at others, alignment can thus occur across different representational dimensions. Recent work (e.g., \citealp{sinclair-etal-2022-structural, jumelet-etal-2024-language}) shows that transformer-based models also exhibit priming effects, though they differ from those observed in human communication. We believe that, these studies may overlook the possibility that such differences may produce qualitatively distinct alignment patterns. In the context of syntactic complexity, humans tend to converge within individual dimensions, whereas LLMs may converge across dimensions. We therefore use linear-CKA to capture these dynamics.

\paragraph{Handling Overlapped Speech:} A key phenomenon distinguishing natural dialogue from monologue and human–LLM interaction is that speakers frequently interrupt, take turns, or backchannel while another speaks, resulting in overlapping speech. We observed that transcription strategies vary across the natural conversation datasets used in this study. For instance, in the BTSJ dataset, backchannels from one speaker are occasionally embedded directly within the other speaker's utterance text. Conversely, in the MUNDEX dataset, overlapping speech is segmented chronologically based on which speaker spoke first. To ensure consistency, we parsed all utterances across all corpora; specifically for the BTSJ dataset, we stripped out backchannels embedded within the primary speaker's turn.

\begin{figure}[t!]
    \centering
    \begin{subfigure}[b]{\columnwidth}
        \centering
        \includegraphics[width=0.75\textwidth]{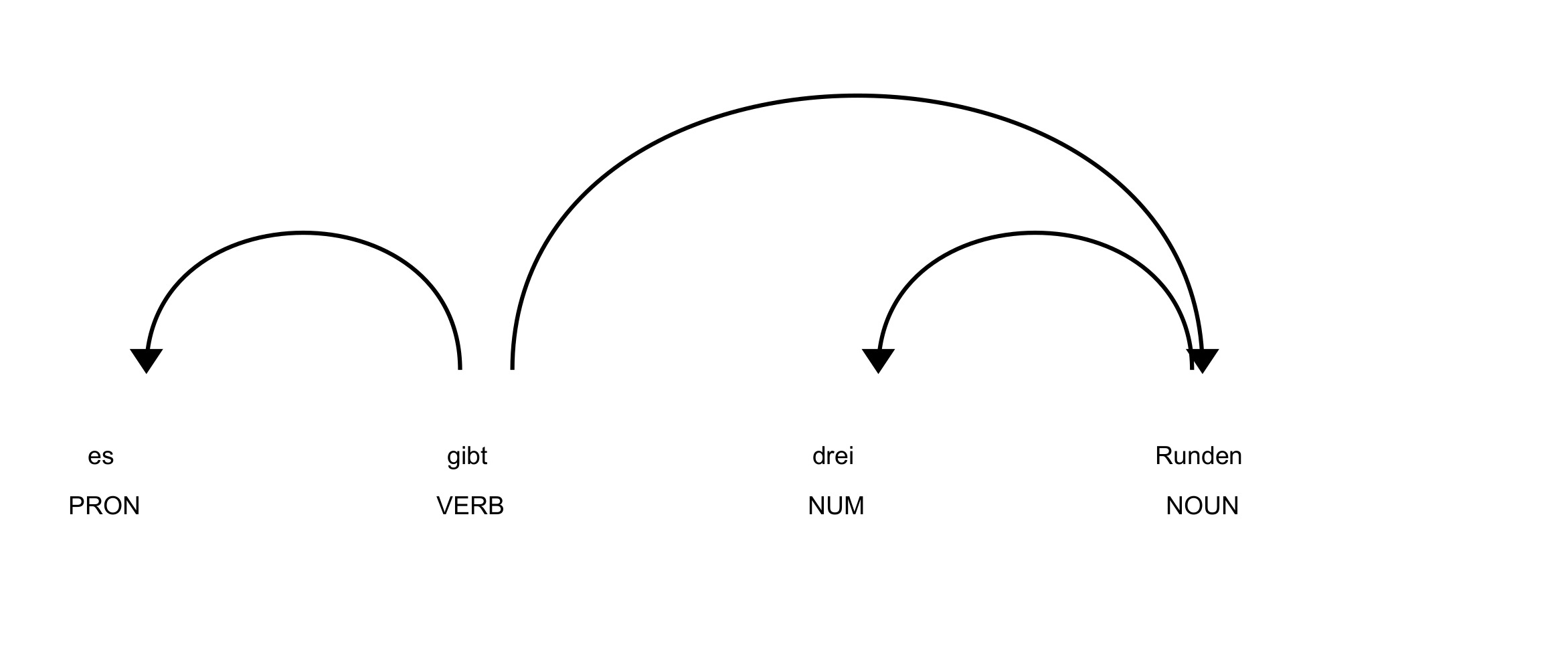}
        \caption{Simple dependency structure (translation: “There are three rounds”), which corresponds to the example shown in Figure~\ref{fig:measure_example_short}.}
         \label{fig:dpsim}
         \vspace{0.5cm}
    \end{subfigure}
    \begin{subfigure}[b]{\columnwidth}
        \centering
        \includegraphics[width=\textwidth]{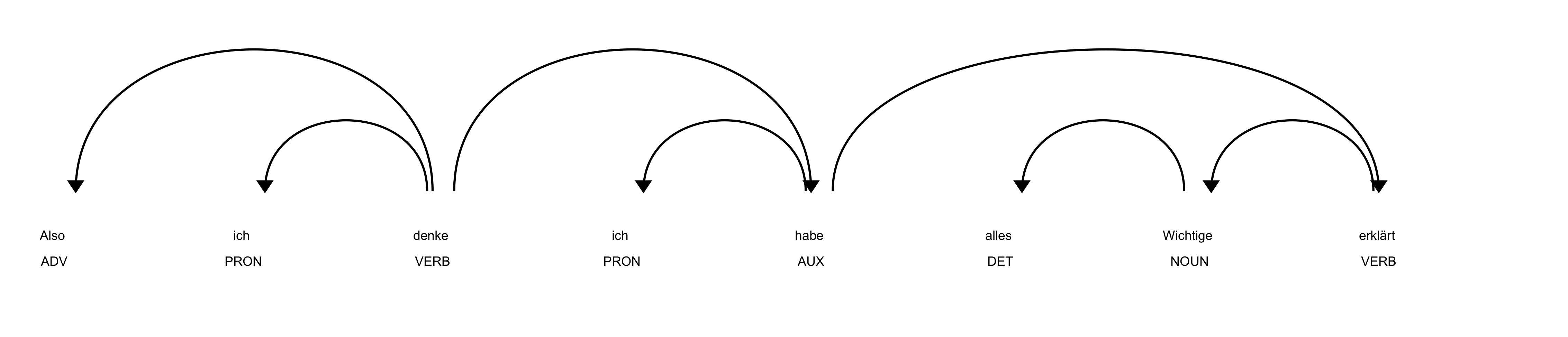}
        \caption{Complex dependency structure (translation: “So I think I have explained all important things”).}
        \label{fig:dpcom}
    \end{subfigure}
    \caption{Two examples showing the syntactic relationships according to UD. Edges are directed from heads to dependents. The root arc is omitted for the simplicity.}
    \label{fig:1df}
\end{figure}

\paragraph{Representation of the Syntactic Complexity:} One fact about the dialogue data we use is that a dialogue consists of many turns, and each turn can contain one or multiple utterances. Some utterances are complete sentences while others are not. In this paper, the measurement is conducted based on turns. Therefore, we use the averages of the syntactic complexity of the utterances to represent the syntactic complexity of each turn. That is to say, for each turn, we first measure how many sentences there are, and based on that, we calculate the syntactic complexity using our proposed metrics. After that, we take the average value from each metric, which is used to represent the syntactic complexity of the utterance in that turn. Here we also add more details by using two examples (dependency-based syntax trees of two sentences) shown in Figure~\ref{fig:1df} to explain how our proposed metric in Section~\ref{sec:syncom_metric} can capture utterances with different degree of syntactic complexity.

\paragraph{Cleaning the Human--LLM Dialogue Data:} For the human--LLM dialogue, we notice that the dataset provided by \citet{zheng2024lmsyschatm} contains a lot of noise. For example, some of the dialogues are about asking the language agents to purify codes, while others are about translating sentences in one language into another. These types of human--LLM conversation are not considered as `dialogue' in this study. We conduct pre-processing and manual checks to exclude such noise.

\paragraph{Further Details of Dialogue:} In Table~\ref{tab:dialogue_statistics}, we report the details of the used dialogues, which include the dialogue type, the number of dialogues in each dataset, and turn information.

\section{Supplementary Plots for Figure~\ref{fig:mean_trend} in in Section~\ref{subsec:sc_mean}}
\label{ap:sup_fig}

\begin{figure*}[h!]
    \begin{subfigure}[b]{1\textwidth}
        \includegraphics[width=\textwidth]{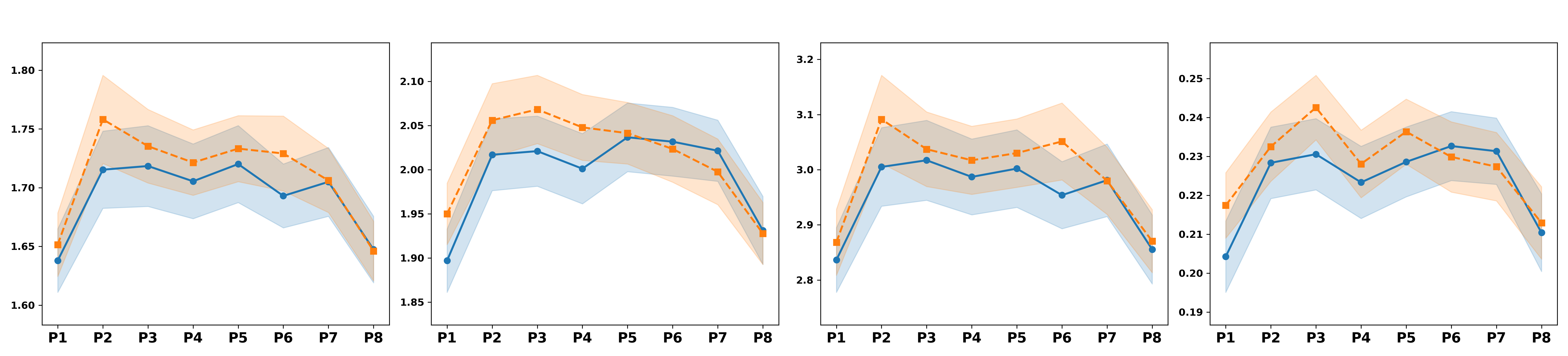}
        \caption{BTSJ Corpus (jpn)}
    \end{subfigure}
    \begin{subfigure}[b]{1\textwidth}
        \includegraphics[width=\textwidth]{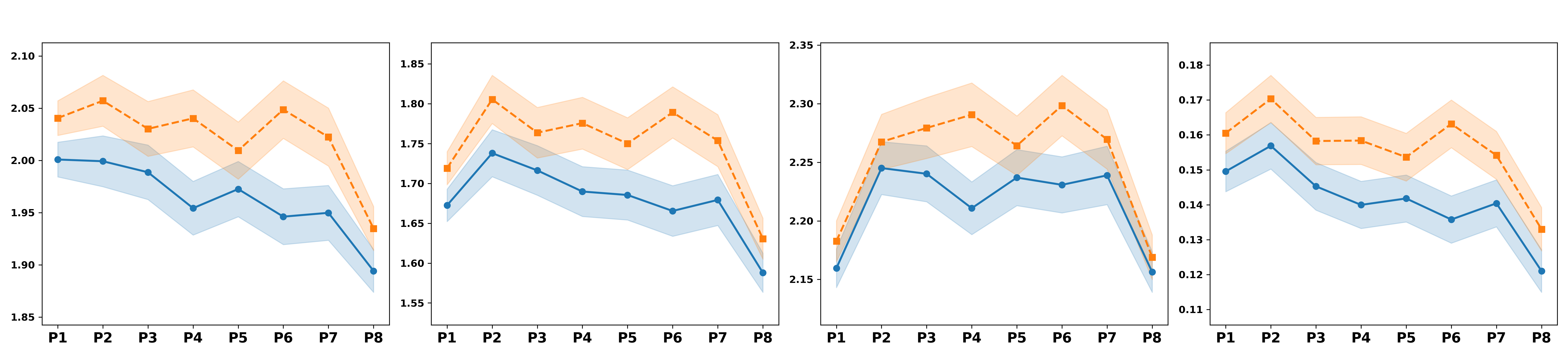}
        \caption{Candor Corpus (eng)}
    \end{subfigure}
    \begin{subfigure}[b]{1\textwidth}
        \includegraphics[width=\textwidth]{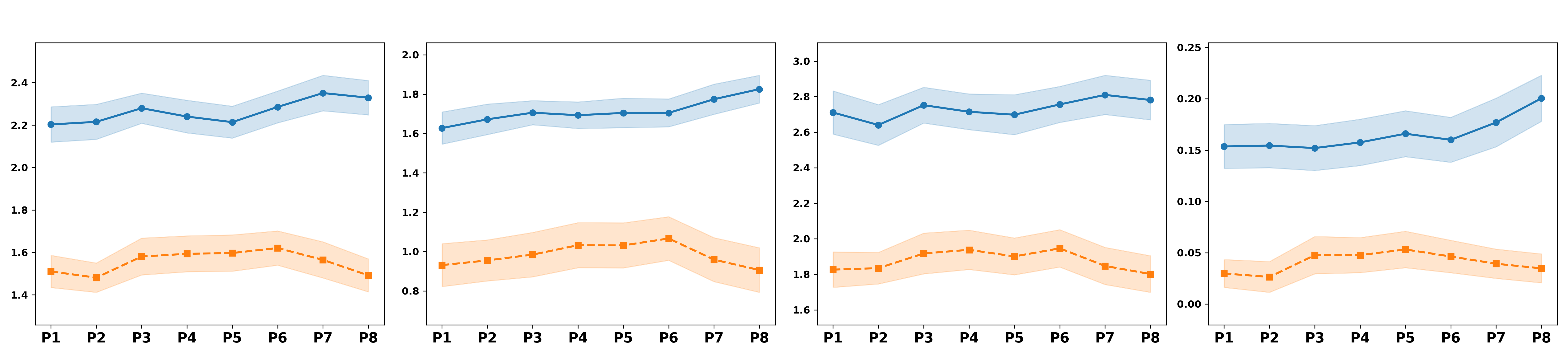}
        \caption{Map Task (eng)}
    \end{subfigure}
    \begin{subfigure}[b]{1\textwidth}
        \includegraphics[width=\textwidth]{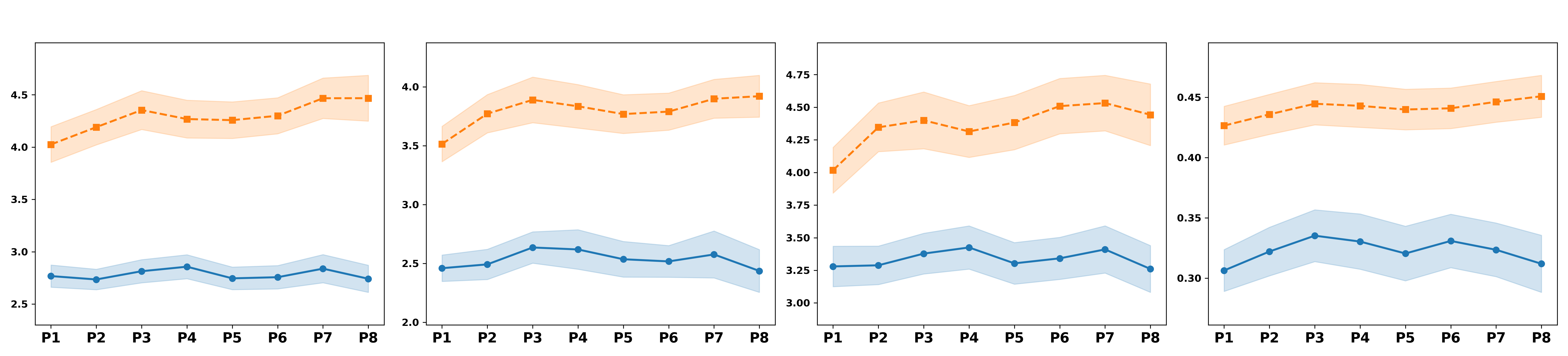}
        \caption{LMSYS-Chat-1M (LLM ger)}
    \end{subfigure}
    \begin{subfigure}[b]{1\textwidth}
        \includegraphics[width=\textwidth]{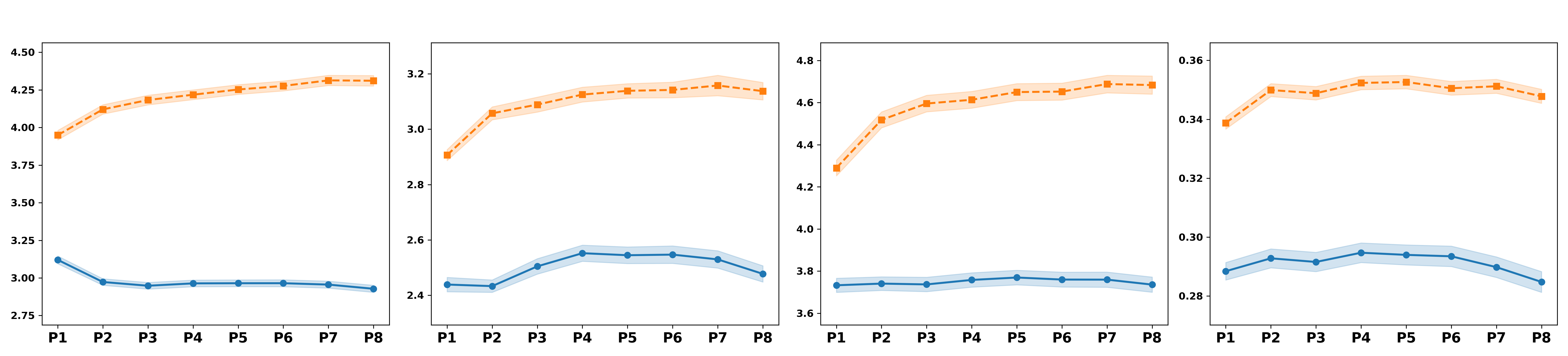}
        \caption{LMSYS-Chat-1M (LLM eng)}
    \end{subfigure}
    \caption{%
        The plots here, as supplimentary plots to Figure~\ref{fig:mean_trend}, show the syntactic complexity across phases for all dialogue datasets except \textit{Mundex} and \textit{Hazumi}. \textbf{Blue line is Dialogue Initiator, Yellow line is Dialogue Follower.} The four metric values in each sub-figure from left to right are: \textbf{syntactic complexity value}, \textbf{dependency length}, \textbf{tree depth}, and \textbf{complex Dependency Structure Ratio}. Syntactic complexity convergence is considered a progressive reduction in the syntactic complexity gap between the two interlocutors over the course of the interaction (from phase 1 to phase 8). We calculate the mean value based on the 8 phases we pre-defined with $95\%$ confidence interval. We conduct pairwise $t$-tests for the value in each phase to make sure that the group of the values from current phase is statistically different from the group of the values from the next phase (except the last phase), the results, reported in Table~\ref{tab:statistical_results} in the Appendix, indicate a strong degree of statistic significance.} 
    \label{fig:mean_trend_sup}
\end{figure*}

Here we put supporting plots in Figure~\ref{fig:mean_trend_sup} for  Figure~\ref{fig:mean_trend} in Section~\ref{subsec:sc_mean} which includes the dialogue datasets showing minor or no syntactic convergence Tendency. We chose only to include plots that show obvious syntactic complexity convergence in the main text, while leaving the others in the Appendix. 

\begin{table*}[ht!]
\scriptsize
\begin{tabularx}{\textwidth}{llllXX}
\toprule
    \textbf{Category} & \textbf{Dataset} & \textbf{Dialogue Type} 
    & \textbf{Number of Dialogue }  & \textbf{Avg. Turns (Init. / Fol.)} & \textbf{Min--Max Turns (Init ./ Fol.)} \\
\midrule
Human-Human & MUNDEX & Task-oriented 
&  87 & 121.07 / 134.62 & 2--426 / 3--771 \\

& HAZUMI & Social/Open 
& 33 & 74.30  / 83.61 & 57--95 / 63--120 \\

& BTSJ & Social/Open
& 505 & 142.78 / 142.77 & 5--484 / 4--485 \\

& CANDOR & Social/Open
& 1855 & 167.91 / 167.93 & 3--917 / 4--917 \\
& MapTask & Task-oriented
& 121 & 166.32  / 130.69 & 24--535 / 18--432 \\

\midrule

Human--LLM& LLM\_GERMAN & Instruction Mixed
& 9849 & 1.95  / 1.95 & 1--45 / 1--45 \\

& LLM\_ENGLISH & Instruction Mixed
& 476737 & 1.65   / 1.65 & 1--215 / 1--215 \\

\bottomrule
\end{tabularx}
\caption{Descriptive Statistics: Dialogue Types and Interaction Metrics per Dataset.Please note that the number of dialogue here is total dialogue number in the dataset, which is different from number of the dialogues used for analysis.}
\label{tab:dialogue_statistics}
\end{table*}

\begin{table*}[h]
\centering
\begin{tabular}{|l|c|c|c|c|}
\hline
\textbf{Dataset} & \textbf{Complexity Score} & \textbf{Dependency Length} & \textbf{Tree Depth} & \textbf{ Dependency Ratio} \\ \hline
Train: Original & 2.9818 & 2.9826 & 4.0773 & 0.3721 \\ \hline
Train: Simplification & 2.9159 & 2.8459 & 3.8692 & 0.3725 \\ \hline
Test: Original & 2.9824 & 2.9887 & 4.0626 & 0.3734 \\ \hline
Test: Simplification & 2.9259 & 2.8621 & 3.8757 & 0.3733 \\ \hline
\end{tabular}
\caption{The average of the syntactic complexity metric values from sentences in train and test set of \textbf{DEplain} corpus. The higher the values, the more syntactically complex the sentences are. Our integrated metrics in general can capture the lower syntactic complexity in the simplified sentences in both training and testing datasets except \textit{Dependency Ratio} in Training dataset.}
\label{tab:syncom_deplain}
\end{table*}

\section{Embedding Syntactic Complexity Features to FourierGPT Pipeline}
\label{sec:fGPT}

In Section~\ref{sec:app} we reported the classification results of using quantified syntactic complexity to detect LLM-involved dialogue. Here we report the technical details on how we integrate the syntactic complexity into the FourierGPT pipeline and use that to perform the classification task. 

In the original FourierGPT, the detection pipeline works as follows: first, raw likelihood scores are estimated by running a forward pass on the input text. Taking an input of $N$ tokens, $t_1,\dots,t_N$, the likelihood score of the $i$th token $s_i$ is computed by $s_i = -\log P(t_i|t_1,\dots,t_{i-1})$, i.e., the surprisal (negative log probability) returned by the language model (also called the \emph{estimator} model). Then, the next step is to obtain the spectrum view of the likelihood sequence $\tilde{s}0,\dots,\tilde{s}{N-1}$ as input. Discrete Fourier transform (\textbf{DFT}) is applied according to the following:

\begin{equation}\label{eq:DFT}
    X(\omega_k)\triangleq \sum_{n=0}^{N-1}\tilde{s}_n e^{-j\omega_k n} .
\end{equation}

The result is a set of complex numbers $\mathcal{F}=\{X(\omega_k)\}|_{k=0,\dots, N-1}$ as the frequency-domain representation of the input time-domain signal (likelihood scores), in which $\omega_k$ is the $k$-th frequency component. 
The starting index of $\tilde{s}_i$ is set to 0 because \textbf{DFT} requires $k=0$ as the lowest frequency component. 
$X(\omega_k)$ is a complex number made up of \textit{real} and \textit{imaginary} parts, $X(\omega_k)=\text{Re}(X(\omega_k))+\text{Im}(X(\omega_k))$. The norm $\lVert X(\omega_k)\rVert=\sqrt{\text{Re}(X(\omega_k))^2 + \text{Im}(X(\omega_k))^2}$ represents the intensity of the $k$th component $\omega_k$.
Finally, the sequence $\{\lVert X(\omega_k)\rVert\}|_{k=0,\dots, N-1}$ is used as the spectrum-view of likelihood, which then provides features for the classification step. The range of $\omega_k$ is $[0,\pi]$. Based on an intuitive interpretation provided by \citet{yang2023face}, the likelihood score $\tilde{s}_i$ at the level of $\lVert X(\omega_k)\rVert$ tends to occur every $1/\omega_k$ tokens in the text data.

In our study, we take syntactic complexity as a linguistic feature which can potentially describe the difference of human-human and human-LLM dialogue. As \textbf{quantified syntactic complexity metric values, similar to surprisal, fluctuate during the development of the dialogue}, it is rational to use \textbf{DFT} to represent the data. However different from the interpretation of periodicity of the likelihood \cite{xu-etal-2024-detecting}, for the syntactic complexity we only assume the fluctuation of the value along the time step. The spectrum view of the four syntactic complexity metric values is thus rather a feature representation we use to capture the fluctuation of the syntactic complexity. 
Formally, for dialogue $d$, we have four syntactic-complexity-score based sequences:
\begin{equation}
\mathbf{s}^{(d,m)}=\{s^{(d,m)}_t\}_{t=0}^{T_{d,m}-1},\quad m\in\{1,2,3,4\},
\end{equation}
Where $d$ denotes dialogue id, $m$ denote metric index, $T$ is the set of the time steps. 
We apply fast Fourier transform to each syntactic complexity metric value sequence and take the power spectrum $p_k$:
\begin{equation}
\begin{aligned}
\hat{s}^{(d,m)}_k&=\sum_{t=0}^{T_{d,m}-1} s^{(d,m)}_t\,e^{-i2\pi kt/T_{d,m}} \\
\quad p^{(d,m)}_k&=\left|\hat{s}^{(d,m)}_k\right|^2
\end{aligned}
\end{equation}

We first compute the spectral representation of each dialogue's syntactic complexity across metric $m$:
\begin{equation}
\mathbf{x}^{(d,m)}=\big[p^{(d,m)}_0,\dots,p^{(d,m)}_{T_{d,m}-1}\big].
\end{equation}
We then concatenate the four spectra into a single feature vector:
\begin{equation}
\mathbf{x}^{(d)}=\mathbf{x}^{(d,1)} \,\|\, \mathbf{x}^{(d,2)} \,\|\, \mathbf{x}^{(d,3)} \,\|\, \mathbf{x}^{(d,4)}.
\end{equation}

Let $\ell_d=|\mathbf{x}^{(d)}|$ denote the length of the feature vector representing dialogue $d$, and let $y_d\in\{0,1\}$ be its binary classification label.

As reported in Table~\ref{tab:dialogue_statistics}, human–LLM dialogues generally contain significantly fewer turns than human–human dialogues. Consequently, a classifier might inadvertently rely on dialogue length as a shortcut feature. To evaluate how effectively quantified syntactic complexity metrics distinguish human–human from human–LLM dialogues compared to surprisal-based values, we introduce a length-balanced sampling strategy. After extracting the \textbf{DFT} spectral values and selecting the top-$k$ frequencies, we construct $B$ quantile bins over the dialogue length distribution $\{\ell_d\}$, ensuring dialogues of similar length fall into the same bin.

For each bin $b$, we define the set of class-specific dialogues as:
\begin{equation}
    \mathcal{D}_{b,c}=\{d:\ell_d\in \text{bin }b,\ y_d=c\},\quad c\in\{0,1\},
\end{equation}
where $c$ represents the ground-truth label ($c=0$ for human–human dialogues and $c=1$ for human–LLM dialogues). To prevent length bias and class imbalance, we uniformly sample without replacement exactly
\begin{equation}
    n_b=\min\{|\mathcal{D}_{b,0}|,\ |\mathcal{D}_{b,1}|\}
\end{equation}
dialogues from each class within bin $b$.

This procedure guarantees that each bin contains an equal number of human–human and human–LLM dialogues of comparable lengths. The resulting balanced dataset is then fed into the training pipeline, forcing the classifiers to rely strictly on syntactic complexity features during evaluation on the test set. To ensure fair comparability, we also apply this same quantile binning strategy to the surprisal-based FourierGPT baseline.

\section{Additional Experiment on Dialogue Length Effect}
\label{ap:length_effect}

One of the primary issues we address here is whether longer dialogues exhibit a higher potential for syntactic complexity convergence than shorter ones. To answer this question, we conducted an experiment in which we incrementally excluded dialogues with fewer than 16, 24, 32, and 36 turns from each dataset. We then recalculated the syntactic complexity convergence ratio following the methodology in Section~\ref{sec:ratio_pattern}; the results are summarized in Tables~\ref{tab:16turns} through \ref{tab:36turns} in the Appendix.

When restricting the analysis to longer interactions by excluding short dialogues, MUNDEX displays an increasing ratio of syntactic complexity convergence. However, length-dependent effects cannot be confirmed for the Hazumi, BTSJ, and Candor datasets under our pre-defined settings, as all dialogues in these three corpora already exceed 36 turns. For human–LLM dialogues, no valid syntactic complexity convergence could be confirmed for longer interactions.

We conducted a similar experiment to test whether syntactic complexity convergence becomes more pronounced across different metrics in dialogues with more turns. The results, summarized in Table~\ref{tab:syncom_dialogue_length} in the Appendix, reveal that after excluding short dialogues, BTSJ and LLM German no longer display significant complexity convergence, whereas MUNDEX maintains a strong effect. For the LLM English dataset, complexity convergence holds under all length constraints. The lack of significant convergence in LLM German is likely attributable to sample size reduction, as very few dialogues remain after filtering (e.g., only 12 dialogues remain after excluding those with fewer than 24 turns), making meaningful correlation analysis unfeasible.

\section{Evaluation of Our Integrated Syntactic Complexity Metrics}
\label{ap:deplain}

\begin{figure*}[t!]
    \centering
    \includegraphics[width=1\linewidth]{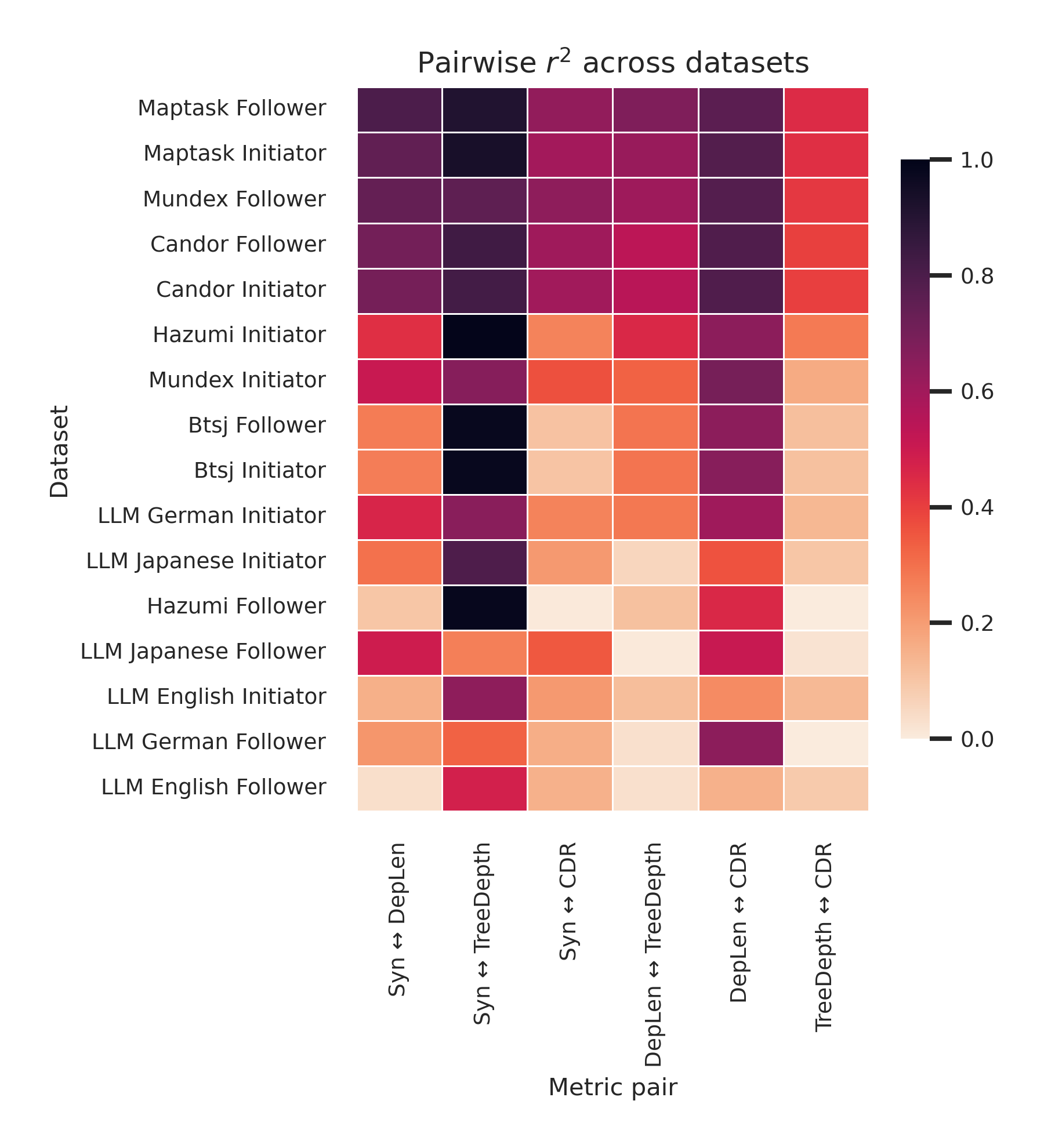}
    \caption{Inter-Correlation of the 4 Integrated Metrics in All Dialogue Datasets. 4 syntactic complexity metrics give us 6 distinct metric pairs.\textit{syn}, \textit{DepLen},  \textit{CDR} denote syntactic complexity score, dependency length and complex dependency ratio.Darker colours indicate strong correlation, lighter colors indicate weak or no correlation.}
    \label{fig:intercor_dialogue}
\end{figure*}

To evaluate the validity of our integrated syntactic complexity metrics, we conducted two experiments. In the first experiment, we calculated the pairwise coefficient of determination ($R^2$) among the four syntactic complexity metrics across all dialogue datasets. Each dataset yields two data points (representing the syntactic complexity of the initiator and follower, respectively). The rationale is that weak or absent correlations between metric pairs indicate that each of the four metrics captures distinct, non-redundant aspects of syntactic complexity. The resulting heat map is shown in Figure~\ref{fig:intercor_dialogue}.

We observed that across all dialogue datasets, Tree Depth and Complex Dependency Ratio consistently exhibit a weak correlation as a metric pair. Furthermore, among the 16 data points, 9 (from \textit{Hazumi Initiator} downward) have more than three metric pairs showing weak correlations. Although Syntactic Complexity Score and Tree Depth generally display a strong correlation, they still show a weak correlation in \textit{LLM Japanese Follower} and \textit{LLM German Follower}. These findings justify using all four syntactic complexity metrics, as no single metric can sufficiently represent the others.

In our second experiment, we evaluated the validity of the integrated metrics on DEplain, an open-source parallel corpus. Its training set contains 10,661 sentence pairs, while its test set contains 1,232 sentence pairs. We applied our integrated syntactic complexity metrics to the dataset to obtain metric values for each sentence pair and calculated the average value for all sentences. As reported in Table~\ref{tab:syncom_deplain}, the simplified sentences generally exhibit lower syntactic complexity metric values, with the exception of the complex dependency ratio in the training set.

\begin{figure*}[t!]
    \centering
    \includegraphics[width=1\linewidth]{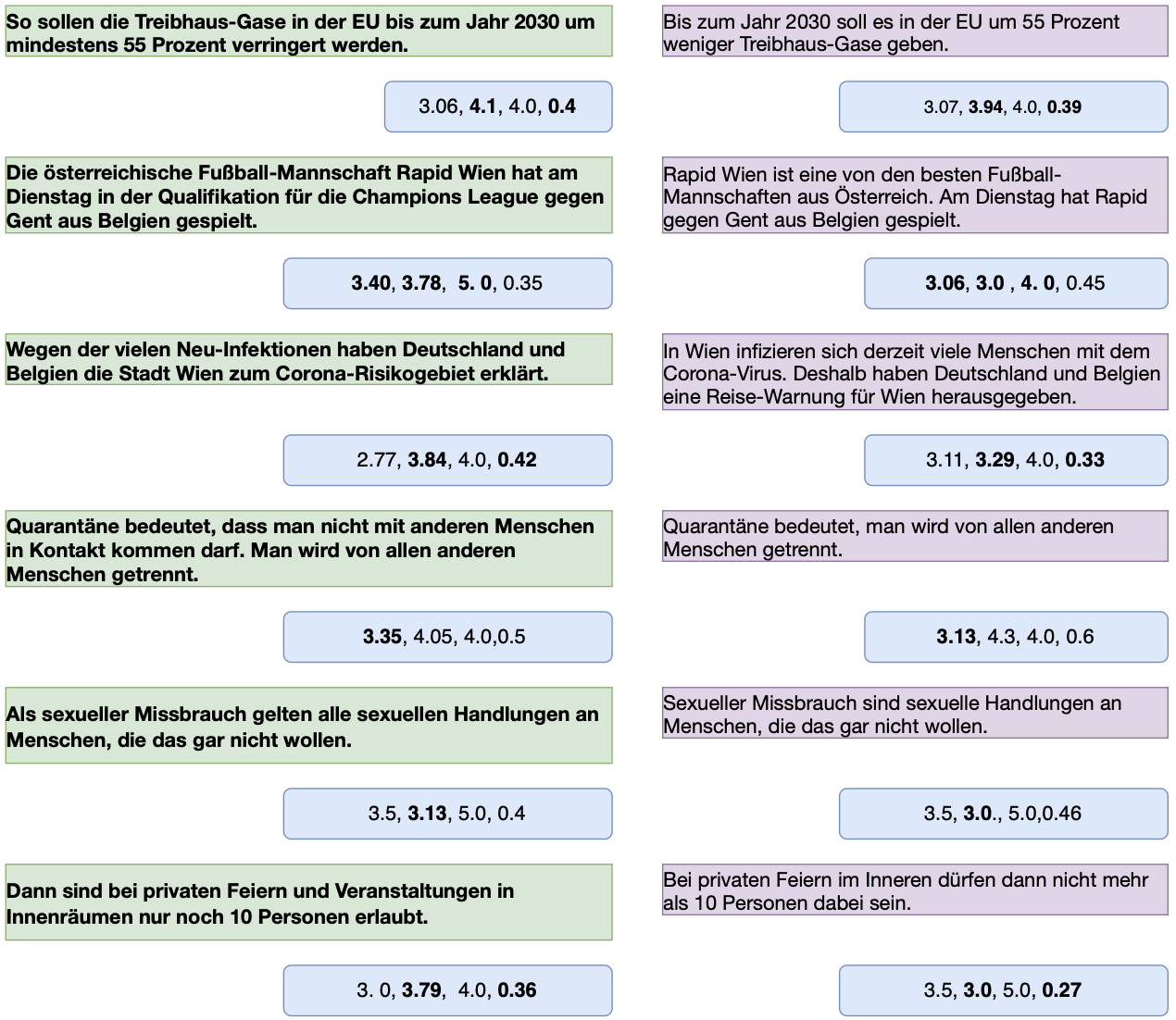}
    \caption{This figure presents six sentence pairs, where the sentences in the green boxes represent the original inputs and those in the purple boxes represent the simplified versions, which we consider syntactically simple. The values in the blue boxes correspond to four syntactic complexity metrics, capturing complexity from distinct linguistic perspectives. Values in bold indicate instances where the change in syntactic complexity was successfully captured by a metric. The values are ordered from left to right as follows: \textbf{Syntactic Complexity Value}, \textbf{Dependency Length}, \textbf{Tree Depth}, \textbf{Complex Dependency Ratio}.}
    \label{fig:sample_syncom_deplain}
\end{figure*}

We acknowledge that in the DEplain dataset, simplified sentences may feature simpler lexical choices rather than structural simplifications. Therefore, for our qualitative analysis, we sampled six sentence pairs where the simplified sentences were specifically selected for their lower syntactic complexity. Applying our proposed metrics to these sentences showed that our four metrics generally capture syntactic complexity effectively, as illustrated in Figure~\ref{fig:sample_syncom_deplain}.

\section{GPU Usage, Parameters Used for Training, and Further Robustness Test for the Classification Shown in Section~\ref{sec:app}}
\label{ap:robust}

To measure syntactic complexity metrics, we used an A6000 64GB GPU to accelerate calculations. The total computation for the human–human dialogues took around 26 hours, whereas the larger human–LLM dataset required approximately 3 days. We also used the GPU to train the transformer-based encoder for human–LLM dialogue detection, which took around 20 minutes. All hyperparameter settings are listed in Table~\ref{tab:parameters} for reference. For classifier training, we sampled 20,000 human–LLM dialogues with varying turn counts from the English subset of LMSYS-Chat-1M. We used the Candor dataset (1,855 dialogues) as the human–human benchmark. Consequently, the majority-class baseline accuracy is $20,000 / 21,855 \approx 91.5\%$. The results in Table~\ref{tab:fft_results} demonstrate that our classifiers successfully learned to distinguish human–human from human–LLM dialogues, with both models achieving over 95.5\% accuracy. Additionally, as reported in Table~\ref{tab:dialogue_statistics}, the average turn count for LLM-involved conversations in the \textit{LLM\_English} dataset is relatively low (1.65 turns). To test the robustness of our syntactic complexity–based classification against turn-count bias, we evaluated whether performance holds after filtering out short human–LLM interactions. Removing dialogues with fewer than 8 turns per participant left 7,850 human–LLM conversations. We then merged these with the human–human dialogues and re-ran the classification task. As shown in Table~\ref{tab:syncom_fft_filter_results}, filtering out short dialogues did not significantly alter classification performance.

\section{The Algorithm Used for Recursive Tree Depth Measurement}
\label{ap:alg1}

\begin{algorithm}[H]
\caption{Dependency Tree Depth Computation}
\begin{algorithmic}[1]
\Require Sentence $S$, dependency parser $P$
\Ensure Tree depth $D$

\State $doc \gets P(S)$
\If{$doc$ has no sentences}
    \State \Return 0
\EndIf

\State $sent \gets$ first sentence in $doc$
\State Initialize empty dictionary $C$ \Comment{$C$ maps head IDs to lists of child IDs}
\State $R \gets$ None \Comment{Root ID}

\For{each word $w$ in $sent$}
    \If{$w.head = 0$}
        \State $R \gets w.id$ \Comment{Found root}
    \Else
        \State Append $w.id$ to $C[w.head]$
    \EndIf
\EndFor

\If{$R$ is None}
    \State \Return 0
\EndIf

\Function{Depth}{$node$}
    \If{$node \notin C$}
        \State \Return 1
    \Else
        \State \Return $1 + \max_{\;child \in C[node]}(\Call{Depth}{child})$
    \EndIf
\EndFunction

\State $D \gets \Call{Depth}{R}$
\State \Return $D$

\end{algorithmic}
\end{algorithm}

\begin{table*}[ht]
\scriptsize
\resizebox{\linewidth}{!}{
\begin{tabular}{lllll}
\toprule
\textbf{Input} & \textbf{Model} & \textbf{F1} & \textbf{Accuracy} & \textbf{AUC} \\
\midrule
\multirow{3}{*}{Syn Com (Original Dialogues)}
 & Spectrum Length (baseline) & 0.745 (0.070) & 0.753 (0.060) & 0.757 (0.063) \\
 & Linear Regression          & 0.962 (0.040) & 0.960 (0.044) & \textbf{1.000 (0.000)} \\
 & Transformer Encoder        & 0.933 (0.067) & 0.927 (0.081) & 0.982 (0.025) \\
\midrule
\multirow{3}{*}{Syn Com (Filtered Dialogues)}
 & Spectrum Length (baseline) & 0.836 (0.054) & 0.832 (0.060) & 0.832 (0.060) \\
 & Linear Regression          & \textbf{0.979 (0.016)} & \textbf{0.978 (0.017)} & 0.999 (0.003) \\
 & Transformer Encoder        & 0.964 (0.030) & 0.962 (0.033) & 0.997 (0.005) \\
\bottomrule
\end{tabular}
}
\caption{Comparison with dialogues all kept against dialogue with more than 8 turns kept}
\label{tab:syncom_fft_filter_results}
\end{table*}

\begin{table*}[t]
\centering
\begin{tabular}{ll}
\toprule
\textbf{Component} & \textbf{Parameters / Values} \\
\midrule
Iterations & $n\_iterations = 10$ \\
LLM sampling & $llm\_sample\_size = 20000$ \\
Length control & \texttt{--length\_match}, $length\_match\_bins = 20$ \\
Feature selection & \texttt{--fourier\_percent} $\in\{100, 50, 25\}$ \\
 & \texttt{--fourier\_top\_k} (optional), \texttt{--fixed\_len} (optional) \\
Normalization & \texttt{--power\_norm} $\in\{\text{none}\}$ \\
Splits (logistic Regression) & \texttt{test\_size} $= 0.3$ (stratified) \\
Splits (Transformer Encoder) & train/val $= 0.8/0.2$, test $= 0.3$ if no ID split \\
Traditional model & StandardScaler; SMOTE (if enabled) \\
Logistic Regression & \texttt{max\_iter} $= 1000$, \texttt{class\_weight} = balanced \\
Transformer encoder & $D = 128$, heads $= 4$, layers $= 2$, dropout $= 0.1$ \\
FFN / classifier & FFN dim $= 256$, classifier hidden $= 64$ \\
Optimizer & Adam, lr $= 10^{-3}$ \\
Scheduler & ReduceLROnPlateau (factor $= 0.5$, patience $= 5$) \\
Training & epochs $= 20$, batch\_size $= 32$ \\
\bottomrule
\end{tabular}
\caption{Training parameters used for, bin sampling and training the logistic regression, and transformer encoder.}
\label{tab:parameters}
\end{table*}

\begin{table*}[t]
\scriptsize
\setlength{\tabcolsep}{5pt}
\centering
\begin{tabular}{llcccc}
\toprule
\textbf{Dataset} & \textbf{Metric} & \textbf{N} & \textbf{Both Down (\%)} & \textbf{Converge (\%)} & \textbf{Total (\%)} \\
\midrule
MUNDEX & Syn Value &  & 34.00 & 42.00 & 76.00 \\
 & Dependency Length &  & 22.00 & 44.00 & 66.00 \\
 & Tree Depth &  & 14.00 & 42.00 & 56.00 \\
 & Complex Dependency Ratio &  & 4.00 & 56.00 & 60.00 \\
\cmidrule(lr){2-6}
 & \textbf{OVERALL (any)} & \textbf{50} & \textbf{38.00} & \textbf{48.00} & \textbf{86.00} \\
\midrule
HAZUMI & Syn Value &  & 9.09 & 42.42 & 51.52 \\
 & Dependency Length &  & 15.15 & 60.61 & 75.76 \\
 & Tree Depth &  & 6.06 & 27.27 & 33.33 \\
 & Complex Dependency Ratio &  & 0.00 & 69.70 & 69.70 \\
\cmidrule(lr){2-6}
 & \textbf{OVERALL (any)} & \textbf{33} & \textbf{21.21} & \textbf{75.76} & \textbf{96.97} \\
\midrule
BTSJ & Syn Value &  & 7.03 & 34.34 & 41.37 \\
 & Dependency Length &  & 10.64 & 38.35 & 49.00 \\
 & Tree Depth &  & 2.61 & 24.30 & 26.91 \\
 & Complex Dependency Ratio &  & 6.83 & 46.39 & 53.21 \\
\cmidrule(lr){2-6}
 & \textbf{OVERALL (any)} & \textbf{498} & \textbf{18.07} & \textbf{57.43} & \textbf{75.50} \\
\midrule
CANDOR & Syn Value &  & 16.71 & 21.99 & 38.71 \\
 & Dependency Length &  & 16.87 & 27.01 & 43.88 \\
 & Tree Depth &  & 1.08 & 20.86 & 21.94 \\
 & Complex Dependency Ratio &  & 12.61 & 33.80 & 46.42 \\
\cmidrule(lr){2-6}
 & \textbf{OVERALL (any)} & \textbf{1855} & \textbf{28.95} & \textbf{40.81} & \textbf{69.76} \\
\midrule
MAPTASK & Syn Value &  & 5.79 & 38.02 & 43.80 \\
 & Dependency Length &  & 4.96 & 37.19 & 42.15 \\
 & Tree Depth &  & 4.13 & 30.58 & 34.71 \\
 & Complex Dependency Ratio &  & 2.48 & 37.19 & 39.67 \\
\cmidrule(lr){2-6}
 & \textbf{OVERALL (any)} & \textbf{121} & \textbf{9.92} & \textbf{50.41} & \textbf{60.33} \\
\midrule
LLM\_GERMAN & Syn Value &  & 15.38 & 27.78 & 43.16 \\
 & Dependency Length &  & 8.97 & 26.92 & 35.90 \\
 & Tree Depth &  & 9.40 & 27.78 & 37.18 \\
 & Complex Dependency Ratio &  & 5.56 & 41.03 & 46.58 \\
\cmidrule(lr){2-6}
 & \textbf{OVERALL (any)} & \textbf{234} & \textbf{23.50} & \textbf{51.28} & \textbf{74.79} \\
\midrule

LLM\_ENGLISH & Syn Value &  & 14.28 & 26.94 & 41.22 \\
 & Dependency Length &  & 8.57 & 32.94 & 41.52 \\
 & Tree Depth &  & 8.18 & 27.48 & 35.66 \\
 & Complex Dependency Ratio &  & 8.08 & 37.03 & 45.11 \\
\cmidrule(lr){2-6}
 & \textbf{OVERALL (any)} & \textbf{7850} & \textbf{27.27} & \textbf{50.00} & \textbf{77.27} \\
\bottomrule
\end{tabular}
\caption{Syntactic complexity convergence: Metric-specific and overall dialogue counts across datasets.}
\label{tab:pattern_ratio_summary}
\end{table*}

\begin{table*}[ht]
\scriptsize
\setlength{\tabcolsep}{5pt}
\centering
\begin{tabular}{lrrllrrrlr}
\toprule
\textbf{Dataset} & \textbf{N} & \textbf{Metric} & \textbf{Obs} & \textbf{Random Mean} & \textbf{NullSD} & \textbf{p-val} & \textbf{Cohen d} & \textbf{CI} \\
\midrule
MUNDEX & 40 & syn value & 0.7750 & 0.4800 & 0.0783 & \textbf{0.0000}$^{***}$ & 3.7649 & \textbf{[0.1999, 0.3897]} \\
 & & dependency length & 0.7000 & 0.4930 & 0.0827 & \textbf{0.0150}$^{*}$ & 2.5046 & \textbf{[0.0996, 0.3147]} \\
 & & tree depth & 0.5500 & 0.2417 & 0.0649 & \textbf{0.0000}$^{***}$ & 4.7541 & \textbf{[0.1930, 0.4250]} \\
 & & complex dependency ratio & 0.6750 & 0.4675 & 0.0770 & \textbf{0.0067}$^{**}$ & 2.6925 & \textbf{[0.0925, 0.3151]} \\ \midrule
HAZUMI & 33 & syn value & 0.5152 & 0.4864 & 0.0829 & 0.4384 & 0.3472 & [-0.0951, 0.1636] \\
 & & dependency length & 0.7576 & 0.4937 & 0.0869 & \textbf{0.0020}$^{**}$ & 3.0355 & \textbf{[0.1563, 0.3758]} \\
 & & tree depth & 0.3333 & 0.3723 & 0.0795 & 0.7495 & -0.4898 & [-0.1617, 0.0823] \\
 & & complex dependency ratio & 0.6970 & 0.4942 & 0.0854 & \textbf{0.0101}$^{*}$ & 2.3762 & \textbf{[0.0940, 0.3392]} \\ \midrule
MAPTASK & 121 & syn value & 0.4380 & 0.4875 & 0.0435 & 0.8987 & -1.1376 & [-0.1140, 0.0188] \\
 & & dependency length & 0.4215 & 0.4896 & 0.0444 & 0.9453 & -1.5342 & [-0.1306, 0.0037] \\
 & & tree depth & 0.3471 & 0.3446 & 0.0439 & 0.5027 & 0.0572 & [-0.0625, 0.0688] \\
 & & complex dependency ratio & 0.3967 & 0.4515 & 0.0443 & 0.9160 & -1.2375 & [-0.1182, 0.0110] \\ \midrule
BTSJ & 476 & syn value & 0.4055 & 0.4696 & 0.0230 & 0.9987 & -2.7841 & [-0.0982, -0.0288] \\
 & & dependency length & 0.4895 & 0.4904 & 0.0226 & 0.5307 & -0.0379 & [-0.0343, 0.0303] \\
 & & tree depth & 0.2584 & 0.2875 & 0.0201 & 0.9427 & -1.4472 & [-0.0595, -0.0003] \\
 & & complex dependency ratio & 0.5315 & 0.4940 & 0.0217 & \textbf{0.0440}$^{*}$ & 1.7287 & \textbf{[0.0044, 0.0713]} \\ \midrule
CANDOR & 1855 & syn value & 0.3871 & 0.4817 & 0.0113 & 1.0000 & -8.4072 & [-0.1109, -0.0793] \\
 & & dependency length & 0.4388 & 0.4828 & 0.0115 & 0.9994 & -3.8377 & [-0.0614, -0.0265] \\
 & & tree depth & 0.2194 & 0.2102 & 0.0090 & 0.1611 & 1.0234 & [-0.0047, 0.0229] \\
 & & complex dependency ratio & 0.4642 & 0.4467 & 0.0114 & \textbf{0.0651}$^{\dagger}$ & 1.5279 & [-0.0019, 0.0358] \\ \midrule
LLM GERMAN & 42 & syn value & 0.5000 & 0.4903 & 0.0790 & 0.5032 & 0.1225 & [-0.1058, 0.1252] \\
 & & dependency length & 0.3095 & 0.4985 & 0.0787 & 0.9952 & -2.4001 & [-0.2985, -0.0819] \\
 & & tree depth & 0.4524 & 0.4044 & 0.0747 & 0.2921 & 0.6426 & [-0.0711, 0.1557] \\
 & & complex dependency ratio & 0.4524 & 0.4908 & 0.0800 & 0.7365 & -0.4804 & [-0.1575, 0.0744] \\ \midrule
LLM ENGLISH & 1313 & syn value & 0.3983 & 0.4921 & 0.0138 & 1.0000 & -6.7929 & [-0.1158, -0.0742] \\
 & & dependency length & 0.4242 & 0.4905 & 0.0137 & 1.0000 & -4.8344 & [-0.0874, -0.0466] \\
 & & tree depth & 0.3618 & 0.3978 & 0.0137 & 0.9960 & -2.6348 & [-0.0550, -0.0171] \\
 & & complex dependency ratio & 0.4593 & 0.4766 & 0.0140 & 0.9000 & -1.2385 & [-0.0396, 0.0034] \\
\bottomrule
\end{tabular}
\caption{Convergence results in dialogues with at least 16 turns for each speaker.}
\label{tab:16turns}
\end{table*}

\begin{table*}[ht]
\scriptsize
\setlength{\tabcolsep}{5pt}
\centering
\begin{tabular}{lrrllrrrlr}
\toprule
\textbf{Dataset} & \textbf{N} & \textbf{Metric} & \textbf{Obs} & \textbf{Random Mean} & \textbf{NullSD} & \textbf{p-val} & \textbf{Cohen d} & \textbf{CI} \\
\midrule
MUNDEX & 37 & syn value & 0.7838 & 0.4850 & 0.0857 & \textbf{0.0000}$^{***}$ & 3.4861 & \textbf{[0.1968, 0.3950]} \\
 & & dependency length & 0.7297 & 0.4917 & 0.0792 & \textbf{0.0036}$^{**}$ & 3.0031 & \textbf{[0.1273, 0.3416]} \\
 & & tree depth & 0.5946 & 0.2284 & 0.0666 & \textbf{0.0000}$^{***}$ & 5.4975 & \textbf{[0.2477, 0.4916]} \\
 & & complex dependency ratio & 0.6757 & 0.4678 & 0.0803 & \textbf{0.0108}$^{*}$ & 2.5873 & \textbf{[0.0922, 0.3322]} \\ \midrule
HAZUMI & 33 & syn value & 0.5152 & 0.4864 & 0.0829 & 0.4384 & 0.3472 & [-0.0951, 0.1636] \\
 & & dependency length & 0.7576 & 0.4937 & 0.0869 & \textbf{0.0020}$^{**}$ & 3.0355 & \textbf{[0.1563, 0.3758]} \\
 & & tree depth & 0.3333 & 0.3723 & 0.0795 & 0.7495 & -0.4898 & [-0.1617, 0.0823] \\
 & & complex dependency ratio & 0.6970 & 0.4942 & 0.0854 & \textbf{0.0101}$^{*}$ & 2.3762 & \textbf{[0.0940, 0.3392]} \\ \midrule
MAPTASK & 120 & syn value & 0.4417 & 0.4860 & 0.0452 & 0.8560 & -0.9793 & [-0.1110, 0.0209] \\
 & & dependency length & 0.4250 & 0.4879 & 0.0455 & 0.9307 & -1.3834 & [-0.1279, -0.0007] \\
 & & tree depth & 0.3417 & 0.3449 & 0.0452 & 0.5720 & -0.0722 & [-0.0672, 0.0661] \\
 & & complex dependency ratio & 0.4000 & 0.4500 & 0.0438 & 0.8960 & -1.1395 & [-0.1166, 0.0158] \\ \midrule
BTSJ & 444 & syn value & 0.4054 & 0.4711 & 0.0224 & 0.9987 & -2.9279 & [-0.1002, -0.0318] \\
 & & dependency length & 0.4752 & 0.4899 & 0.0242 & 0.7480 & -0.6046 & [-0.0481, 0.0191] \\
 & & tree depth & 0.2500 & 0.2827 & 0.0209 & 0.9467 & -1.5633 & [-0.0641, -0.0027] \\
 & & complex dependency ratio & 0.5293 & 0.4908 & 0.0228 & \textbf{0.0520}$^{\dagger}$ & 1.6861 & \textbf{[0.0022, 0.0755]} \\ \midrule
CANDOR & 1855 & syn value & 0.3871 & 0.4817 & 0.0113 & 1.0000 & -8.4072 & [-0.1109, -0.0793] \\
 & & dependency length & 0.4388 & 0.4828 & 0.0115 & 0.9994 & -3.8377 & [-0.0614, -0.0265] \\
 & & tree depth & 0.2194 & 0.2102 & 0.0090 & 0.1611 & 1.0234 & [-0.0047, 0.0229] \\
 & & complex dependency ratio & 0.4642 & 0.4467 & 0.0114 & \textbf{0.0651}$^{\dagger}$ & 1.5279 & [-0.0019, 0.0358] \\ \midrule
LLM GERMAN & 12 & syn value & 0.5000 & 0.4864 & 0.1350 & 0.5767 & 0.1008 & [-0.2007, 0.2636] \\
 & & dependency length & 0.4167 & 0.4842 & 0.1486 & 0.7700 & -0.4544 & [-0.3175, 0.1408] \\
 & & tree depth & 0.5000 & 0.3819 & 0.1350 & 0.2700 & 0.8744 & [-0.0962, 0.3323] \\
 & & complex dependency ratio & 0.5000 & 0.4853 & 0.1588 & 0.5700 & 0.0927 & [-0.1996, 0.2290] \\ \midrule
LLM ENGLISH & 428 & syn value & 0.3551 & 0.4819 & 0.0236 & 1.0000 & -5.3727 & [-0.1607, -0.0909] \\
 & & dependency length & 0.4089 & 0.4830 & 0.0244 & 1.0000 & -3.0372 & [-0.1114, -0.0398] \\
 & & tree depth & 0.3411 & 0.3702 & 0.0231 & 0.9093 & -1.2577 & [-0.0625, 0.0029] \\
 & & complex dependency ratio & 0.4720 & 0.4714 & 0.0243 & 0.4987 & 0.0240 & [-0.0344, 0.0363] \\
\bottomrule
\end{tabular}
\caption{Convergence results in dialogues with at least 24 turns for each speaker.}
\label{tab:24turns}
\end{table*}

\begin{table*}[ht]
\scriptsize
\setlength{\tabcolsep}{5pt}
\centering
\begin{tabular}{lrrllrrrlr}
\toprule
\textbf{Dataset} & \textbf{N} & \textbf{Metric} & \textbf{Obs} & \textbf{Random Mean} & \textbf{NullSD} & \textbf{p-val} & \textbf{Cohen d} & \textbf{CI} \\
\midrule
MUNDEX & 33 & syn value & 0.8485 & 0.4903 & 0.0863 & \textbf{0.0000}$^{***}$ & 4.1482 & \textbf{[0.2716, 0.4597]} \\
 & & dependency length & 0.7879 & 0.5011 & 0.0871 & \textbf{0.0000}$^{***}$ & 3.2930 & \textbf{[0.1831, 0.3989]} \\
 & & tree depth & 0.6061 & 0.2198 & 0.0697 & \textbf{0.0000}$^{***}$ & 5.5422 & \textbf{[0.2564, 0.5302]} \\
 & & complex dependency ratio & 0.6667 & 0.4694 & 0.0858 & \textbf{0.0162}$^{*}$ & 2.3004 & \textbf{[0.0761, 0.3306]} \\ \midrule
HAZUMI & 33 & syn value & 0.5152 & 0.4864 & 0.0829 & 0.4384 & 0.3472 & [-0.0951, 0.1636] \\
 & & dependency length & 0.7576 & 0.4937 & 0.0869 & \textbf{0.0020}$^{**}$ & 3.0355 & \textbf{[0.1563, 0.3758]} \\
 & & tree depth & 0.3333 & 0.3723 & 0.0795 & 0.7495 & -0.4898 & [-0.1617, 0.0823] \\
 & & complex dependency ratio & 0.6970 & 0.4942 & 0.0854 & \textbf{0.0101}$^{*}$ & 2.3762 & \textbf{[0.0940, 0.3392]} \\ \midrule
MAPTASK & 119 & syn value & 0.4454 & 0.4840 & 0.0464 & 0.8213 & -0.8321 & [-0.1090, 0.0292] \\
 & & dependency length & 0.4286 & 0.4876 & 0.0473 & 0.9027 & -1.2474 & [-0.1253, 0.0055] \\
 & & tree depth & 0.3445 & 0.3425 & 0.0415 & 0.5253 & 0.0502 & [-0.0625, 0.0685] \\
 & & complex dependency ratio & 0.4034 & 0.4496 & 0.0433 & 0.8867 & -1.0679 & [-0.1118, 0.0170] \\ \midrule
BTSJ & 419 & syn value & 0.3986 & 0.4711 & 0.0234 & 1.0000 & -3.1045 & [-0.1089, -0.0393] \\
 & & dependency length & 0.4558 & 0.4930 & 0.0244 & 0.9347 & -1.5241 & [-0.0719, -0.0019] \\
 & & tree depth & 0.2411 & 0.2769 & 0.0199 & 0.9707 & -1.8050 & [-0.0668, -0.0049] \\
 & & complex dependency ratio & 0.5155 & 0.4932 & 0.0242 & 0.1800 & 0.9210 & [-0.0158, 0.0611] \\ \midrule
CANDOR & 1855 & syn value & 0.3871 & 0.4817 & 0.0113 & 1.0000 & -8.4072 & [-0.1109, -0.0793] \\
 & & dependency length & 0.4388 & 0.4828 & 0.0115 & 0.9994 & -3.8377 & [-0.0614, -0.0265] \\
 & & tree depth & 0.2194 & 0.2102 & 0.0090 & 0.1611 & 1.0234 & [-0.0047, 0.0229] \\
 & & complex dependency ratio & 0.4642 & 0.4467 & 0.0114 & \textbf{0.0651}$^{\dagger}$ & 1.5279 & [-0.0019, 0.0358] \\ \midrule
LLM GERMAN & 5 & syn value & 0.4000 & 0.4893 & 0.2170 & 0.8100 & -0.4116 & [-0.4893, 0.1773] \\
 & & dependency length & 0.4000 & 0.5053 & 0.2239 & 0.8267 & -0.4704 & [-0.5053, 0.1613] \\
 & & tree depth & 0.4000 & 0.3713 & 0.2203 & 0.5933 & 0.1301 & [-0.3713, 0.2953] \\
 & & complex dependency ratio & 0.4000 & 0.5340 & 0.2226 & 0.8267 & -0.6020 & [-0.5340, 0.1327] \\ \midrule
LLM ENGLISH & 184 & syn value & 0.3533 & 0.4888 & 0.0375 & 1.0000 & -3.6128 & [-0.1888, -0.0870] \\
 & & dependency length & 0.4239 & 0.4910 & 0.0358 & 0.9720 & -1.8731 & [-0.1204, -0.0148] \\
 & & tree depth & 0.3370 & 0.3711 & 0.0354 & 0.8627 & -0.9626 & [-0.0841, 0.0199] \\
 & & complex dependency ratio & 0.4728 & 0.4835 & 0.0355 & 0.6240 & -0.2997 & [-0.0661, 0.0391] \\
\bottomrule
\end{tabular}
\caption{Convergence results in dialogues with at least 32 turns for each speaker.}
\label{tab:32turns}
\end{table*}

\begin{table*}[ht]
\scriptsize
\setlength{\tabcolsep}{5pt}
\centering
\begin{tabular}{lrrllrrrlr}
\toprule
\textbf{Dataset} & \textbf{N} & \textbf{Metric} & \textbf{Obs} & \textbf{Random Mean} & \textbf{NullSD} & \textbf{p-val} & \textbf{Cohen d} & \textbf{CI} \\
\midrule
MUNDEX & 32 & syn value & 0.8438 & 0.4866 & 0.0872 & \textbf{0.0000}$^{***}$ & 4.0981 & \textbf{[0.2634, 0.4634]} \\
 & & dependency length & 0.7812 & 0.5018 & 0.0870 & \textbf{0.0021}$^{**}$ & 3.2113 & \textbf{[0.1801, 0.3931]} \\
 & & tree depth & 0.5938 & 0.2150 & 0.0709 & \textbf{0.0000}$^{***}$ & 5.3408 & \textbf{[0.2555, 0.5122]} \\
 & & complex dependency ratio & 0.6562 & 0.4731 & 0.0885 & \textbf{0.0312}$^{*}$ & 2.0688 & \textbf{[0.0562, 0.3164]} \\ \midrule
HAZUMI & 33 & syn value & 0.5152 & 0.4864 & 0.0829 & 0.4384 & 0.3472 & [-0.0951, 0.1636] \\
 & & dependency length & 0.7576 & 0.4937 & 0.0869 & \textbf{0.0020}$^{**}$ & 3.0355 & \textbf{[0.1563, 0.3758]} \\
 & & tree depth & 0.3333 & 0.3723 & 0.0795 & 0.7495 & -0.4898 & [-0.1617, 0.0823] \\
 & & complex dependency ratio & 0.6970 & 0.4942 & 0.0854 & \textbf{0.0101}$^{*}$ & 2.3762 & \textbf{[0.0940, 0.3392]} \\ \midrule
MAPTASK & 118 & syn value & 0.4407 & 0.4858 & 0.0420 & 0.8720 & -1.0747 & [-0.1092, 0.0277] \\
 & & dependency length & 0.4237 & 0.4878 & 0.0445 & 0.9373 & -1.4423 & [-0.1242, 0.0051] \\
 & & tree depth & 0.3390 & 0.3437 & 0.0413 & 0.5960 & -0.1139 & [-0.0653, 0.0617] \\
 & & complex dependency ratio & 0.4068 & 0.4512 & 0.0441 & 0.8693 & -1.0072 & [-0.1051, 0.0163] \\ \midrule
BTSJ & 408 & syn value & 0.3897 & 0.4714 & 0.0252 & 1.0000 & -3.2349 & [-0.1139, -0.0428] \\
 & & dependency length & 0.4485 & 0.4932 & 0.0245 & 0.9733 & -1.8254 & [-0.0785, -0.0067] \\
 & & tree depth & 0.2353 & 0.2733 & 0.0215 & 0.9680 & -1.7652 & [-0.0677, -0.0045] \\
 & & complex dependency ratio & 0.5049 & 0.4931 & 0.0245 & 0.3253 & 0.4837 & [-0.0256, 0.0509] \\ \midrule
CANDOR & 1855 & syn value & 0.3871 & 0.4817 & 0.0113 & 1.0000 & -8.4072 & [-0.1109, -0.0793] \\
 & & dependency length & 0.4388 & 0.4828 & 0.0115 & 0.9994 & -3.8377 & [-0.0614, -0.0265] \\
 & & tree depth & 0.2194 & 0.2102 & 0.0090 & 0.1611 & 1.0234 & [-0.0047, 0.0229] \\
 & & complex dependency ratio & 0.4642 & 0.4467 & 0.0114 & \textbf{0.0651}$^{\dagger}$ & 1.5279 & [-0.0019, 0.0358] \\ \midrule
LLM GERMAN & 5 & syn value & 0.4000 & 0.4893 & 0.2170 & 0.8100 & -0.4116 & [-0.4893, 0.1773] \\
 & & dependency length & 0.4000 & 0.5053 & 0.2239 & 0.8267 & -0.4704 & [-0.5053, 0.1613] \\
 & & tree depth & 0.4000 & 0.3713 & 0.2203 & 0.5933 & 0.1301 & [-0.3713, 0.2953] \\
 & & complex dependency ratio & 0.4000 & 0.5340 & 0.2226 & 0.8267 & -0.6020 & [-0.5340, 0.1327] \\ \midrule
LLM ENGLISH & 134 & syn value & 0.3507 & 0.4884 & 0.0431 & 1.0000 & -3.1923 & [-0.1977, -0.0781] \\
 & & dependency length & 0.4478 & 0.4885 & 0.0451 & 0.8373 & -0.9032 & [-0.1029, 0.0234] \\
 & & tree depth & 0.3284 & 0.3680 & 0.0411 & 0.8520 & -0.9633 & [-0.1022, 0.0187] \\
 & & complex dependency ratio & 0.4552 & 0.4813 & 0.0427 & 0.7493 & -0.6111 & [-0.0931, 0.0425] \\
\bottomrule
\end{tabular}
\caption{Convergence results in dialogues with at least 36 turns for each speaker.}
\label{tab:36turns}
\end{table*}

\begin{table*}[ht]
\footnotesize
\setlength{\tabcolsep}{4pt}
\centering
\begin{tabularx}{\textwidth}{llXXXX}
\toprule
\textbf{Dataset} & \textbf{Role} & \textbf{syn\_value} & \textbf{dep\_length} & \textbf{tree\_depth} & \textbf{complex\_dep} \\
\midrule
\multirow{2}{*}{MUNDEX} 
    & Initiator & $150.33^{***}$ & $116.78^{***}$ & $111.79^{***}$ & $61.09^{***}$ \\
    & Follower  & $11.98^{ns}$   & $10.84^{ns}$   & $7.14^{ns}$    & $15.02^{*}$ \\
\midrule
\multirow{2}{*}{btsj}
    & Initiator & $35.24^{***}$ & $89.81^{***}$ & $38.83^{***}$ & $62.15^{***}$ \\
    & Follower  & $51.62^{***}$ & $70.94^{***}$ & $51.73^{***}$ & $67.24^{***}$ \\
\midrule
\multirow{2}{*}{candor}
    & Initiator & $121.19^{***}$ & $84.53^{***}$ & $67.98^{***}$ & $105.23^{***}$ \\
    & Follower  & $110.36^{***}$ & $91.66^{***}$ & $101.82^{***}$ & $114.61^{***}$ \\
\midrule
\multirow{2}{*}{Hazumi}
    & Initiator & $24.58^{***}$ & $19.61^{**}$ & $25.08^{***}$ & $21.42^{**}$ \\
    & Follower  & $41.61^{***}$ & $30.22^{***}$ & $33.23^{***}$ & $5.47^{ns}$ \\
\midrule
\multirow{2}{*}{maptask}
    & Initiator & $14.25^{*}$ & $16.68^{*}$ & $5.36^{ns}$ & $17.97^{*}$ \\
    & Follower  & $10.87^{ns}$ & $9.08^{ns}$ & $7.27^{ns}$ & $12.35^{ns}$ \\
\midrule
\multirow{2}{*}{llm\_ger}
    & Initiator & $2.79^{ns}$ & $8.96^{ns}$ & $4.69^{ns}$ & $8.11^{ns}$ \\
    & Follower  & $21.65^{**}$ & $48.17^{***}$ & $22.57^{**}$ & $19.24^{**}$ \\
\midrule
\multirow{2}{*}{llm\_eng}
    & Initiator & $96.55^{***}$ & $60.48^{***}$ & $6.35^{ns}$ & $38.40^{***}$ \\
    & Follower  & $837.49^{***}$ & $759.82^{***}$ & $438.42^{***}$ & $139.33^{***}$ \\
\bottomrule
\multicolumn{6}{l}{\small Significance levels: $^{***} p < 0.001$, $^{**} p < 0.01$, $^{*} p < 0.05$, $^{ns}$ not significant.}
\end{tabularx}
\caption{Friedman test results across datasets for initiator and follower roles across all of selected datasets.}
\label{tab:statistical_results}
\end{table*}

\begin{table*}[t]
\footnotesize
\centering
\begin{tabularx}{\linewidth}{lXXXX}
\toprule
\textbf{Dataset} & \textbf{MinTurns>=16} & \textbf{MinTurns>=24} & \textbf{MinTurns>=32} & \textbf{MinTurns>=36} \\
\midrule
MUNDEX
& 0.0038 (0.111, 0.420) 
& 0.0032 (0.129, 0.383) 
& \textbf{0.0061 (0.596, 0.025)*} 
& \textbf{0.0059 (0.678, 0.012)*} \\

HAZUMI
& 0.0075 (0.170, 0.310) 
& 0.0075 (0.170, 0.310) 
& 0.0075 (0.170, 0.310) 
& 0.0075 (0.170, 0.310) \\

BTSJ
& \textbf{-0.0036 (0.875, 0.001)***}
& \textbf{-0.0018 (0.604, 0.023)*}
& -0.0004 (0.079, 0.499)
& -0.0000 (0.000, 0.990) \\

MAPTASK
& -0.0011 (0.164, 0.319)
& -0.0010 (0.084, 0.486)†
& -0.0013 (0.133, 0.375)
& -0.0011 (0.082, 0.491)† \\

CANDOR
& -0.0004 (0.005, 0.872)
& -0.0004 (0.005, 0.872)
& -0.0004 (0.005, 0.872)
& -0.0004 (0.005, 0.872) \\

LLM\_ENGLISH
& \textbf{-0.0150 (0.916, 0.000)***}
& \textbf{-0.0121 (0.963, 0.000)***}
& \textbf{-0.0101 (0.844, 0.001)***}
& \textbf{-0.0087 (0.567, 0.031)*} \\

LLM\_GERMAN
& -0.0172 (0.437, 0.074)†
& -0.0093 (0.100, 0.446)
& -0.0087 (0.039, 0.638)
& -0.0087 (0.039, 0.638) \\
\bottomrule
\end{tabularx}
\caption{The syntactic complexity convergence across different metrics with longer dialogue length. The value outside of the parenthesis is the slope value. Parentheses show (correlation, p-value). Significance markers: † $p<.10$, * $p<.05$, ** $p<.01$, *** $p<.001$.}
\label{tab:syncom_dialogue_length}
\end{table*}

\end{document}